\documentclass[conference]{IEEEtran}
\usepackage{amsmath,amsfonts}
\usepackage{algorithmic}
\usepackage{algorithm}
\usepackage{array}
\usepackage[caption=false,font=normalsize,labelfont=sf,textfont=sf]{subfig}
\usepackage{textcomp}
\usepackage{stfloats}
\usepackage{cuted}  
\usepackage{url}
\usepackage{verbatim}
\usepackage{graphicx}
\usepackage{cite}
\usepackage{subfiles}
\usepackage{booktabs}
\usepackage{xcolor}
\usepackage{multirow}
\usepackage{wrapfig}
\usepackage{colortbl}
\usepackage{makecell}
\usepackage{pifont}

\usepackage{xspace}
\usepackage{bm}
\usepackage{amssymb}
\usepackage{hyperref}
\definecolor{hyperblue}{rgb}{0,0,1}
\hypersetup{colorlinks,linkcolor=hyperblue,citecolor=hyperblue,urlcolor=hyperblue}
\graphicspath{{figs/}}
\newcommand{\ourmethod}{Uni-LaViRA\xspace}

\renewcommand{\thetable}{\Roman{table}}

\usepackage{iftex}
\ifPDFTeX\else\usepackage{lmodern}\fi

\makeatletter
\AtBeginDocument{%
  \long\def\@makecaption#1#2{%
    \def\@IEEEcaptype@tmp{table}%
    \ifx\@captype\@IEEEcaptype@tmp
      {\normalfont\footnotesize\centering {#1.}\nobreakspace\nobreakspace #2\par}%
      \@IEEEtablecaptionsepspace
      \vskip 4pt
    \else
      \@IEEEfigurecaptionsepspace
      \setbox\@tempboxa\hbox{\normalfont\footnotesize {#1.}\nobreakspace\nobreakspace #2}%
      \ifdim \wd\@tempboxa >\hsize
        \setbox\@tempboxa\hbox{\normalfont\footnotesize {#1.}\nobreakspace\nobreakspace}%
        \parbox[t]{\hsize}{\normalfont\footnotesize\noindent\unhbox\@tempboxa #2}%
      \else
        \ifCLASSOPTIONconference \hbox to\hsize{\normalfont\footnotesize\hfil\box\@tempboxa\hfil}%
        \else \hbox to\hsize{\normalfont\footnotesize\box\@tempboxa\hfil}\fi
      \fi
    \fi}%
}
\makeatother

\begin{document}

\title{Uni-LaViRA: Language-Vision-Robot Actions Translation for Unified
Embodied Navigation}

\author{%
  \IEEEauthorblockN{%
    Hongyu~Ding\textsuperscript{1,2,*},
    Sizhuo~Zhang\textsuperscript{3,*},
    Ziming~Xu\textsuperscript{1,*},
    Jinwen~Guo\textsuperscript{1},
    Hongxiu~Liu\textsuperscript{1},
    Xingzhi~Cheng\textsuperscript{1},\\
    Zixuan~Chen\textsuperscript{1},
    Haifei~Qi\textsuperscript{4},
    Duo~Wang\textsuperscript{4},
    Hao~Xu\textsuperscript{1},
    Jieqi~Shi\textsuperscript{1,\textdagger},
    Yifan~Zhang\textsuperscript{2,\textdagger},\\
    Jing~Huo\textsuperscript{1,\textdagger},
    Jian~Cheng\textsuperscript{2},
    Yang~Gao\textsuperscript{1},
    and~Jiebo~Luo\textsuperscript{5}%
  }
  \vspace{8pt}
  \IEEEauthorblockA{%
    \textsuperscript{1}Nanjing University, Nanjing, China \quad
    \textsuperscript{3}Beihang University, Beijing, China\\
    \textsuperscript{2}Institute of Automation, Chinese Academy of Sciences, Beijing, China\\
    \textsuperscript{4}BMW (Nanjing) Information Technology Co., Ltd., Nanjing, China\\
    \textsuperscript{5}University of Rochester, Rochester, NY, USA%
  }%
  \vspace{6.3pt}
  \IEEEauthorblockA{%
    \normalsize Project page: \url{https://xetroubadour.github.io/Uni-LaViRA/}%
  }%
}

\maketitle

\setlength\stripsep{-15pt}
\begin{strip}
  {\centering
   \includegraphics[width=\textwidth]{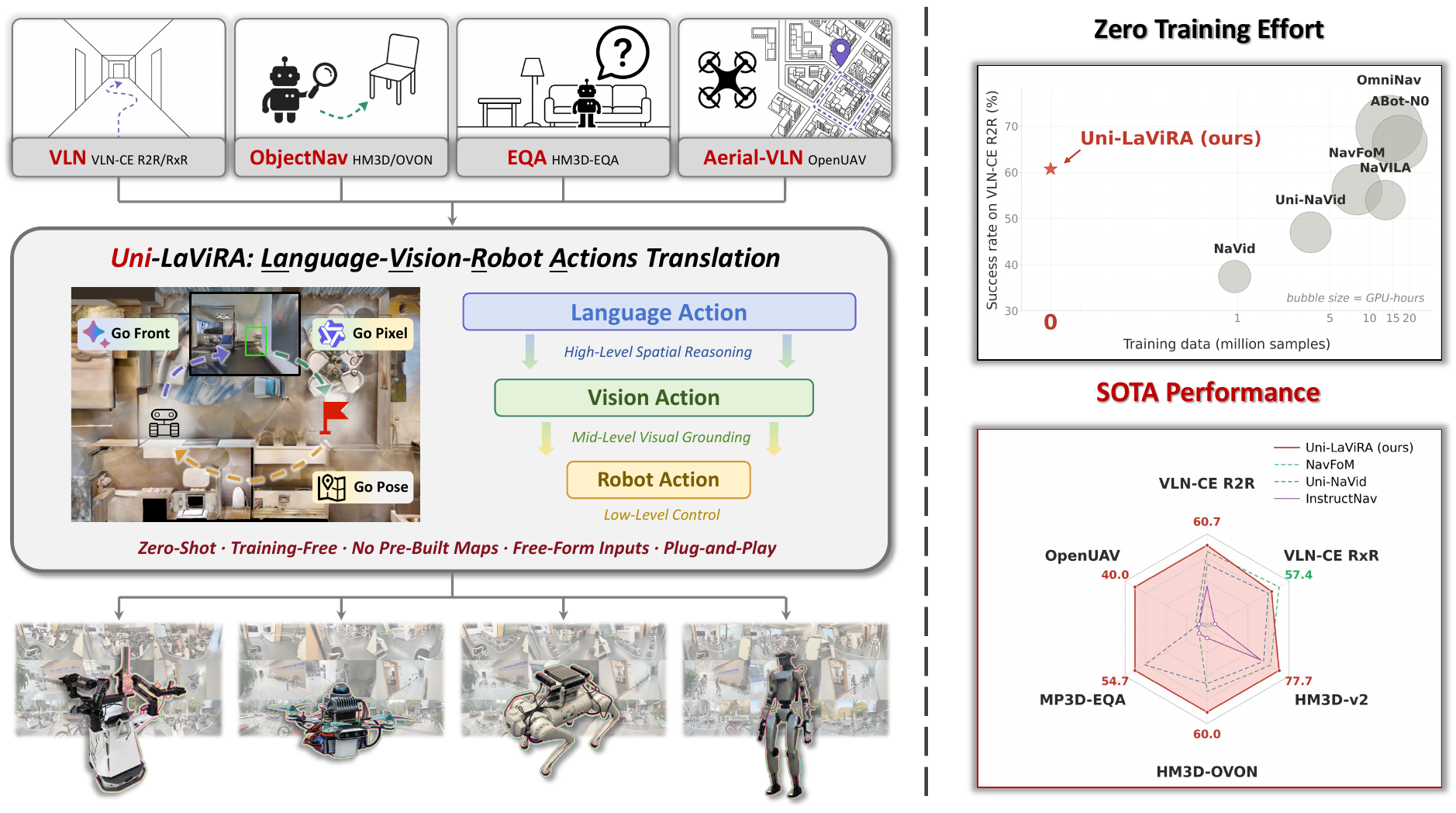}\par}
  \vspace{3pt}
  \refstepcounter{figure}\label{fig:teaser}
  {\footnotesize Fig.~\thefigure.\quad \textbf{\ourmethod in one picture.}
  (Left) A single zero-shot agentic architecture realises
  Language\,$\rightarrow$\,Vision\,$\rightarrow$\,Robot
  Action translation across four task families and four real
  embodiments.  (Right) With zero training effort, \ourmethod reaches
  state-of-the-art performance against training navigation foundation
  models that consume millions of samples and thousands of
  GPU-hours.\par}
  \vspace{17.5pt}
\end{strip}

\begin{abstract}
Embodied navigation requires an agent to map language and visual observations
to a stream of spatial actions that drive a real robot through environments it
has never seen. The dominant approach has been to scale vision-language-action
(VLA) foundation models on ever-larger collections of robot trajectories. This
paper argues that, for navigation specifically, generality can be obtained
structurally, not only through data scale. The underlying decision structure
of navigation reduces to a single \emph{Language-Vision-Robot Actions
Translation}. The language action emits semantic-level directional command and
the vision action emits a pixel-level visual target. Both outputs lie inside
the natural output manifold of pretrained multimodal large language models
(MLLMs), so the task can be reasoned about by an \emph{agent} rather than
learned from robot data. Therefore, we present \textbf{\ourmethod}, a unified
agentic architecture that extends the same insight to four task families
(VLN-CE, ObjectNav, EQA, and Aerial-VLN) and to four heterogeneous real robots
(Wheeled, Quadruped, Humanoid robot, and a self-built UAV) in a zero-shot
manner. Two agent-loop mechanisms make this unification practical. TODO List
Memory (TDM) rewrites a structured checklist of pending sub-goals at every
step, reciting the unfinished items back into the agent's most recent
attention window. Second Chance Backtrack (SCB) rolls the robot back to the
pre-error state and conditions the agent's next plan on the failed
sub-trajectory, turning single-pass navigation into a self-correcting process.
With zero training effort, \ourmethod reaches 60.7\% SR on VLN-CE R2R, 51.3\%
on VLN-CE RxR, 77.7\% on HM3D-v2, 60.0\% on HM3D-OVON, 54.7\% on MP3D-EQA, and
40.00\% on OpenUAV, matching or even surpassing recent training navigation
foundation models that consume millions of samples and thousands of GPU-hours.
\end{abstract}


\section{Introduction}
\label{sec:introduction}

Embodied navigation, the problem of moving a physical
agent through a previously unseen environment in response to a
language instruction, sits at the intersection of computer vision,
natural language understanding, and robotics.  It has emerged as
one of the central testbeds for general-purpose embodied
intelligence.
Its task families have expanded over the past decade. Vision-and-language
navigation (VLN-CE)~\cite{anderson2018vision,krantz2020beyond,ku2020rxr}
requires an agent to follow step-by-step natural-language instructions through
indoor scenes. Object-goal navigation
(ObjectNav)~\cite{batra2020objectnav,ramakrishnan2021hm3d,yokoyama2024hm3dovon}
asks the agent to locate an instance of a named object category in an
unfamiliar environment. Embodied question answering
(EQA)~\cite{das2018eqa,majumdar2024openeqa} couples navigation with active
perception, requiring the agent to explore until it can answer a question
grounded in the scene. Aerial vision-and-language navigation
(Aerial-VLN)~\cite{wang2024openuav} extends instruction following to outdoor,
three-dimensional flight.  These tasks differ in instruction format, sensing
modality, and embodiment, but share a common substrate.  The agent must
perceive a visual scene, interpret a language input, and emit a sequence of
spatial actions that move it toward an intended state.  {Building a single
system that generalises across this spectrum is a long-standing goal of the
field.}%

Over the past two years, the dominant answer to generality in embodied
navigation has converged on a single recipe, training ever-larger
vision-language-action (VLA) foundation models on ever-growing collections of
robot trajectories~\cite{zhang2024navid,
zhang2024uninavid,cheng2024navila,wei2025streamvln,zhang2025navfom}. Each
generation grows training data and compute by orders of magnitude and reports
commensurate gains on performance and task coverage, in effect aiming at a
scaling law for embodied navigation. This line of work stakes cross-task and
cross-embodiment generalisation on data scale. {This paper argues for a
complementary source.} For navigation specifically, generality can be obtained
structurally rather than through data scale alone, once the task is decomposed
in the right way. Across VLN-CE, ObjectNav, EQA, and Aerial-VLN, the
underlying decision structure is stable across all four. The agent reasons
about the instruction, grounds it in the current observation, and emits a
spatial action. {Our conference work LaViRA~\cite{ding2025lavira} made the
central observation that this structure can be realised as a
\emph{Language-Vision-Robot Actions Translation}. The language action emits
semantic-level directional command and the vision action emits a pixel-level
visual target. Both outputs coincide with representations that modern
multimodal large language models (MLLMs) encounter repeatedly during
pretraining.} Every level therefore lands inside what we call the
\emph{natural output manifold} of foundation models, and the task can be
reasoned about by an agent rather than learned by a policy trained on robot
data. {This observation has, we believe, received less attention than it
warrants from a community largely focused on scaling end-to-end navigation
VLAs.}

Our position can be framed as a simple question that motivates this work: does
a task's action space fall inside or outside the pretrained manifold of modern
MLLMs? Within embodied AI, this splits tasks into two regimes, those that are
contact-rich and those that are mostly contact-free. Long-horizon dexterous
manipulation sits firmly on the contact-rich side. Its action semantics, such
as joint torques, contact forces, and impedance schedules, emerge from
continuous physical interaction with objects. They appear as raw numerical
sequences that MLLMs rarely encounter during pretraining. As a result, they
sit firmly outside the MLLM manifold, and such tasks still require end-to-end
VLA learning~\cite{black2024pi0,kim2024openvla}. Mainstream navigation methods
today, by contrast, are mostly contact-free, although cluttered scenes can
still involve incidental collisions or light interaction. A navigating agent
does not negotiate forces with the world but moves through it, and its
decisions reduce naturally to spatial reasoning in language, vision, and
coordinate space. {This physical property is precisely what makes the LaViRA
decomposition possible, and it is what allows navigation, alone among embodied
behaviours, to fully inherit the generalisation power of pretrained MLLMs
without any additional training.}

Beyond the unified framework itself, \ourmethod introduces two mechanisms that
address recurring failure modes of online agentic navigation. The first,
\emph{TODO List Memory} (TDM), reflects a key lesson from recent practice in
long-horizon agent
design~\cite{yao2023react,shinn2023reflexion,huang2022inner}. A long-horizon
agent benefits less from a longer context than from a better-attended one.
Rather than feeding the agent a growing conversation history at every step,
TDM maintains an explicit, dynamically updated list of the sub-tasks implied
by the instruction, {recording what has been completed, what remains, and what
is in progress}. This serves two purposes. It forces the agent to commit to a
high-level plan in a structured form before it emits the next language-level
action, turning each step into an explicit plan-then-act decision rather than
an implicit one. And by rewriting the list at every decision step, TDM
\emph{recites} the unfinished sub-goals back into the agent's most recent
attention window. The second mechanism, Second Chance Backtrack (SCB),
revisits how navigation agents should handle their own mistakes. Backtracking
itself is not new. SmartWay~\cite{shi2025smartway} introduced a backtrack
action for VLN-CE, and LaViRA~\cite{ding2025lavira} used a basic action-level reversal. Both treat
an erroneous step as something to be undone and discarded. SCB takes the
opposite stance. When the agent detects a recent decision that has led the
trajectory off-instruction, it rolls the robot back to the pre-error spatial
state and conditions the next plan on the failed sub-trajectory. The agent
then re-plans with explicit awareness of what did not work. Treating errors as
informative rather than as noise turns single-pass agentic navigation into a
self-correcting process.

This paper is a systematic extension of our preliminary conference version
LaViRA~\cite{ding2025lavira}, which validated the Language--Vision--Robot
Actions Translation idea on a single task family (VLN-CE). LaViRA established
the three-level agentic spine, but three limitations of the conference version
become bottlenecks as soon as one tries to apply the same architecture beyond
instruction-following VLN. \textbf{(L1) Single-task scope.} The conference
prompt schemas, action set, and stopping logic were specialised to
step-by-step instruction following. A single architecture covering ObjectNav,
EQA, and Aerial-VLN under one prompt interface had not been demonstrated.
\textbf{(L2) Attention drift on long instructions.} On RxR, where mean
instruction length is roughly four times that of R2R at about $120$ words, and
on multi-stage UAV flight plans, the LaViRA Language Action Model frequently
forgot which intermediate sub-goals had been satisfied. The agent then emitted
\texttt{stop} prematurely after the first matched landmark. \textbf{(L3) Error
reasoning after a wrong decision.} LaViRA included a primitive reversal action
that returned the agent to the previous waypoint, but the failed
sub-trajectory was discarded, so recovery was effectively a blind retry.

Relative to the conference version, the present manuscript extends the work
along the following four axes.

\begin{itemize}
\item \textbf{From single-task to unified embodied navigation
(addressing~L1).}
We extend the three-level agentic framework from VLN-CE alone to
four heterogeneous task families, namely VLN-CE,
ObjectNav, EQA, and Aerial-VLN, under a
single prompt interface, action set, and controller stack, with
the Language Action Model and Vision Action
Model reused verbatim across tasks.
\item \textbf{TODO List Memory (TDM, addressing~L2).}
We introduce a structured working-memory mechanism that maintains
the agent's progress as an explicit, dynamically updated list of
pending and completed sub-tasks.  TDM
externalises the agent's plan into a form that is re-read at every
decision step, so long multi-clause instructions remain fully
tracked across long horizons.
\item \textbf{Second Chance Backtrack (SCB, addressing~L3).}
We extend LaViRA's primitive reversal action into a full
error-aware re-planning mechanism.  Erroneous trajectories are not
erased but presented back to the agent as reasoning context after
the embodiment is restored to its pre-error state.  This turns
single-pass agentic navigation into a self-correcting process.
\item \textbf{Cross-embodiment real-world deployment and extensive
new analyses.}
The same agentic core is deployed on four heterogeneous real
robots, a wheeled bimanual Agilex Cobot~Magic, a Unitree~G1
humanoid, a Unitree~Go1 quadruped, and a self-built quadrotor UAV,
by swapping only the low-level controller.  We additionally
provide per-task TDM/SCB ablations, a $1{,}800$-trial failure-mode
taxonomy, and an inference-cost analysis.
\end{itemize}

\begin{table}[!t]
\centering
\caption{Scope of LaViRA and \ourmethod. The latter generalises the
agentic core from one task family on two embodiments to four families on four
heterogeneous robots, and adds two new agent-loop mechanisms.}
\label{tab:lavira-vs-unilavira}
\footnotesize
\setlength{\tabcolsep}{3pt}
\renewcommand{\arraystretch}{1.1}
\resizebox{0.85\columnwidth}{!}{%
\begin{tabular}{@{}c c c@{}}
\toprule
\textbf{Aspect} & \textbf{LaViRA} & \textbf{\ourmethod} \\
\midrule
Task families     & 1               & 4 \\
Real embodiments  & 2               & 4 \\
Working memory    & Prompt history  & TDM \\
Error recovery    & 1-step revert   & SCB \\
Failure analysis  & Qualitative     & Quantitative + scaling \\
\bottomrule
\end{tabular}%
}%

\end{table}

Table~\ref{tab:lavira-vs-unilavira} summarises the scope and methodological
insights distinguishing the two versions. The remainder of this paper is
organised as follows. Section~\ref{sec:related} reviews related work on
vision-and-language navigation, zero-shot navigation pipelines, navigation
foundation models, and agentic AI systems. Section~\ref{sec:method} presents
the \ourmethod framework, including the unified problem definition, the
agentic architecture, the TODO List Memory and Second Chance Backtrack
mechanisms, and the unified inference procedure. Section~\ref{sec:experiments}
reports zero-shot results across the four task families, validates the
100-episode subset, compares against trained foundation models, and provides
ablation studies, an inference-cost analysis, and a failure-mode taxonomy.
Section~\ref{sec:realworld} reports real-robot deployments across the four
heterogeneous platforms, and Section~\ref{sec:conclusion} concludes.

\section{Related Work}
\label{sec:related}

\subsection{Vision-and-Language Navigation}

VLN was introduced by Anderson~et~al.~\cite{anderson2018vision} in the
discrete Matterport3D graph setting~\cite{chang2017matterport3d} and later
extended to continuous environments (VLN-CE) by
Krantz~et~al.~\cite{krantz2020beyond}, with multilingual dense instructions
introduced in RxR~\cite{ku2020rxr}. Learned methods have progressed steadily
through cross-modal alignment~\cite{hong2021vlnbert,chen2022duet}, explicit
map representations~\cite{wang2023gridmm,an2023bevbert, an2024etpnav},
large-scale data augmentation~\cite{wang2023scalevln}, and curriculum
learning~\cite{zhu2020babywalk}. The strongest end-to-end supervised methods on VLN-CE R2R are recent
foundation models such as OmniNav~\cite{xue2025omninav} and
ABot-N0~\cite{chen2026abotn0}, which reach roughly $66$--$70\%$ SR by training
on heterogeneous robot data; we discuss this regime in detail in
Section~\ref{subsec:nav_foundation_models}. All of these methods rely on
environment-specific training. The learned weights encode scene priors from
the training distribution and do not transfer to new environments, task
variations, or embodiments without costly retraining. \ourmethod matches and
in several cases surpasses these methods with zero robot data, indicating that
scene-specific learning is not required when the action space already lies
inside the natural output manifold of pretrained MLLMs.

\subsection{Zero-Shot Navigation}

Zero-shot methods avoid robot-data training but introduce other dependencies.
\emph{Waypoint-based} approaches pair an LLM with a pretrained waypoint
predictor~\cite{qiao2024opennav,shi2025smartway,hong2022bridging}. The
predictor proposes discrete candidates and the LLM selects among them, but the
pretrained predictor is itself environment-specific and its candidate set
bounds where the agent can go. \emph{Value-mapping} approaches generate a
semantic heatmap from a VLM and navigate toward the
peak~\cite{chen2025canav,long2024instructnav,yokoyama2024vlfm,
yin2025gcvln,zhou2023esc,kuang2024openfmnav}. These exploit strong VLMs for
open-vocabulary grounding, but typically use the LLM only \emph{offline} to
parse the instruction. A third line treats LLMs as planners over discrete
topological graphs~\cite{zhou2024navgpt,chen2024mapgpt,long2024discussnav}, a
formulation that does not extend naturally to continuous control. Lifelong,
multi-modal-goal systems such as GOAT~\cite{chang2024goat,khanna2024goatbench}
contribute instance-aware semantic memory but still rely on trained skills for
navigation primitives. {The conference version of this work,
LaViRA~\cite{ding2025lavira}, established the three-level agentic
decomposition on VLN-CE. The present paper extends it to four task families
and adds the two agent-loop mechanisms developed in
Sections~\ref{subsec:tdm}--\ref{subsec:scb}.} \ourmethod requires no waypoint
predictor, because the Vision Action Model grounds directly on raw pixels, and
keeps the Language Action Model reasoning online at every decision step in
continuous environments.

\subsection{Navigation Foundation Models}
\label{subsec:nav_foundation_models}

The most prominent contemporary alternative is to train a single VLA model on
heterogeneous robot data and let scale do the work. Starting from VLM-based
navigation on monocular video~\cite{zhang2024navid}, subsequent foundation
models pushed along orthogonal axes of unified task
coverage~\cite{zhang2024uninavid,xue2025omninav}, action-space tokenisation
for new embodiments~\cite{cheng2024navila}, streaming
inference~\cite{wei2025streamvln}, memory
architectures~\cite{zeng2025janusvln,wei2025internvlan1}, and large-scale
cross-embodiment training~\cite{zhang2025navfom,chen2026abotn0}. Across
roughly two years, training data has grown from below $1$\,M to over $16$\,M
trajectories spanning quadrupeds, drones, wheeled robots, humanoids, and
vehicles, while reported SR on VLN-CE R2R has climbed from below $40\%$ to
roughly $70\%$. The sequence reads as a coherent attempt to demonstrate a
\emph{scaling law for embodied navigation}: each generation grows training
data, compute, and task coverage by orders of magnitude, with reported
performance climbing in step. {\ourmethod takes the opposite stance. Because
the decision structure of navigation already lies inside the natural output
manifold of pretrained MLLMs, no training above the low-level controller is
required, and each upper-level agent inherits MLLM backbone improvements
directly without consuming additional robot data.}

\subsection{Agentic AI in the Physical World}

The rise of agentic AI, in which foundation models reason, plan, and act
through structured interaction with tools and
environments~\cite{yao2023react}, has reshaped thinking about AI capabilities
across domains. Representative systems range from SayCan~\cite{ahn2022saycan}
and Code-as-Policies~\cite{liang2023cap} to Voyager~\cite{wang2024voyager},
Inner Monologue~\cite{huang2022inner}, Reflexion~\cite{shinn2023reflexion},
and Agent-S~\cite{agashe2025agents}. Hierarchical decomposition has a long
history in reinforcement learning~\cite{bellman1966dynamic,sutton1999between}
and has appeared in VLN as feudal planners~\cite{liu2023azhp,zhao2025nava3}.

\textbf{Memory representations for long-horizon agents.}
MemGPT~\cite{packer2023memgpt} treats the context window as an OS-managed
virtual memory with explicit paging between a fast working set and an external
store; Generative Agents~\cite{park2023generative} maintain a time-stamped
memory stream with periodic reflection that summarises recent events into
higher-level beliefs; MemoryBank~\cite{zhong2024memorybank} adds long-term
retention with an Ebbinghaus-style forgetting schedule; and
Voyager~\cite{wang2024voyager} grows a structured skill library as procedural
memory for embodied agents. These representations are designed for
general-purpose conversational or open-ended agents and store mostly free-form
items; long-horizon embodied navigation specifically needs to track which
sub-goals of a multi-clause instruction are still pending and which have
already been satisfied at a specific scene observation. The
TDM mechanism we propose in this paper sits in the same lineage but is
tailored to navigation: it maintains an ordered, verifiable checklist of
sub-goals that the agent re-reads and updates at every step, so completion is
grounded in observations rather than left to free-form recall.

\textbf{Error recovery in navigation.} SmartWay~\cite{shi2025smartway} and
LaViRA~\cite{ding2025lavira} both treat a failed step as
something to be undone and discarded. Discarding the failed sub-trajectory,
however, also discards the signal of \emph{why} the previous decision was
wrong, so the agent re-decides from the same prior with no new evidence and
frequently repeats the same mistake. SCB, the second mechanism
we introduce, instead conditions the re-decision on the failed
sub-trajectory itself, turning error evidence into planning context. This is
closer in spirit to the self-critique pattern of
Reflexion~\cite{shinn2023reflexion} but operates at the spatial-trajectory
level rather than the text-trial level.

\section{The \ourmethod Framework}
\label{sec:method}

\textbf{Difference from LaViRA.} Our method builds upon the three-level
Language-Vision-Robot pipeline originally introduced in
LaViRA~\cite{ding2025lavira}. In the present version, we extend it by scaling
from one task family and two embodiments to four task families and four
heterogeneous embodiments, and by adding two agent-loop mechanisms, TODO List
Memory (Section~\ref{subsec:tdm}) and Second Chance Backtrack
(Section~\ref{subsec:scb}). This extension is necessary because the conference
version's prompts and action set were specialised to instruction following,
its working memory degraded on long-horizon directives, and its blind-retry
recovery left repeated errors uncorrected, all of which become bottlenecks
once the pipeline is applied to ObjectNav, EQA, and Aerial-VLN.

Section~\ref{subsec:problem} abstracts the four task families into a single
decision problem. Section~\ref{subsec:architecture} instantiates the
three-level spine under the resulting unified interface.
Sections~\ref{subsec:tdm} and~\ref{subsec:scb} introduce TODO List Memory and
Second Chance Backtrack, the two new agent-loop mechanisms that this paper
adds to the LaViRA spine. Section~\ref{subsec:algorithm} assembles them into a
single per-episode loop.

\begin{figure*}[!t]
  \centering
  \includegraphics[width=0.85\linewidth]{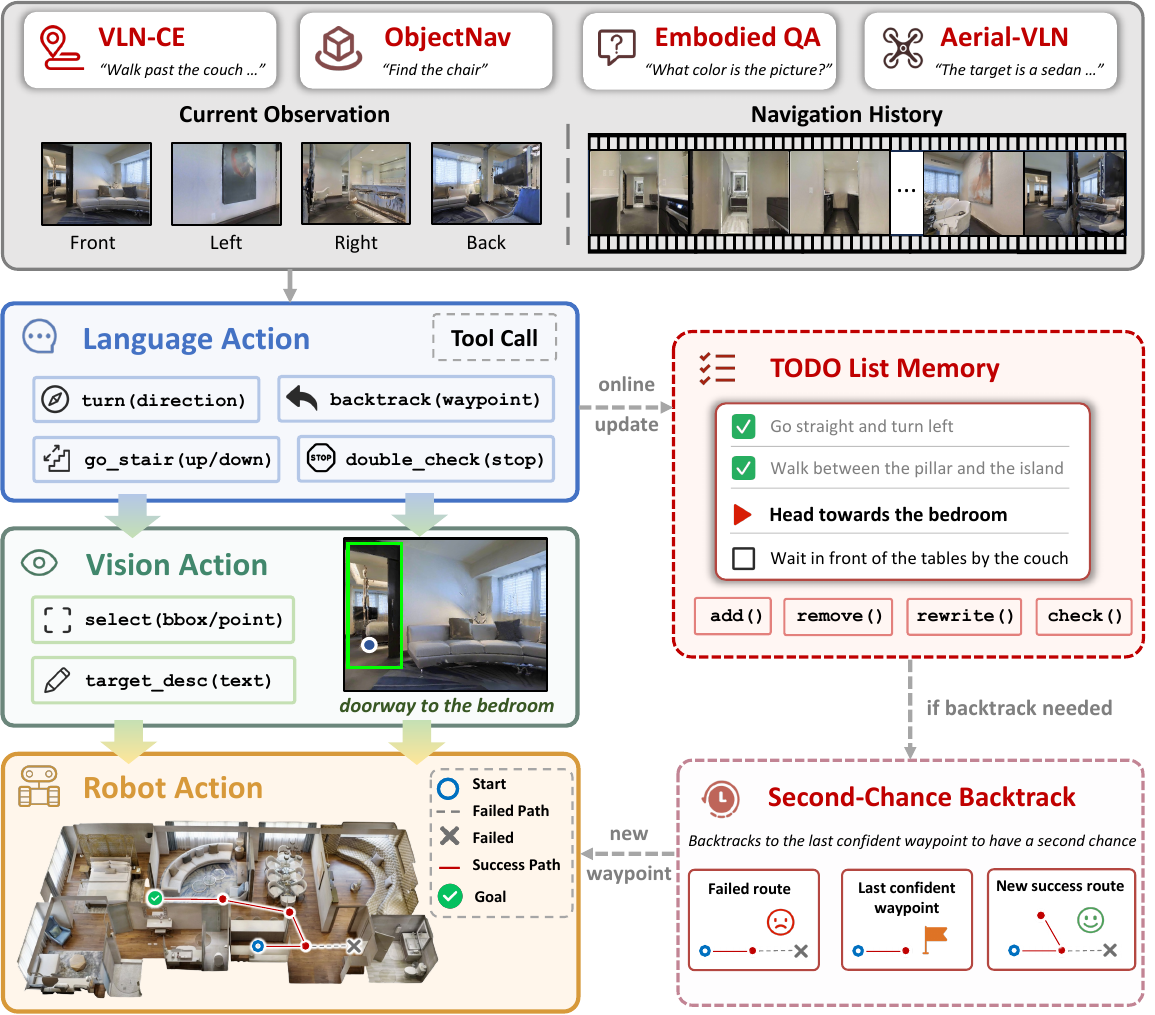}
  \caption{\textbf{The \ourmethod pipeline.}  Language Action emits a
  tool call, Vision Action grounds it on the agent's first-person view
  along the chosen direction, and Robot Action dispatches it.  TDM
  maintains an online checklist; SCB rewinds to a prior waypoint when
  a sub-goal fails. Together, the three-level decomposition
  and the two agent-loop mechanisms turn each step into a verifiable,
  self-correcting decision shared by all four task families.}
  \label{fig:pipeline}
\end{figure*}

\subsection{Problem Definition}
\label{subsec:problem}

We consider four families of embodied navigation. The first three run inside
Habitat-Sim~\cite{krantz2020beyond}. Vision-and-language navigation in
continuous environments (VLN-CE)~\cite{anderson2018vision,ku2020rxr} covers
R2R and RxR. Object-goal navigation (ObjectNav) covers HM3D-v2 and HM3D-OVON.
Embodied question answering (EQA) uses
MP3D-EQA~\cite{das2018eqa,majumdar2024openeqa}. Aerial vision-and-language
navigation (Aerial-VLN) uses OpenUAV~\cite{wang2024openuav} inside AirSim.
Despite differing simulators, embodiments, and goal specifications, all four
families share the same input and output interface from the policy's perspective.

\textbf{Unified interface.} At every decision step $t$, an agent receives
\begin{itemize}
  \item a task specification $\mathcal{T}$ expressed in natural
  language.  This is a route description for
  VLN-CE, an object category or open-vocabulary phrase for
  ObjectNav, a question for EQA, or a flight directive for Aerial-VLN.
  \item an egocentric observation $\mathcal{O}_t$ consisting of RGB and
  aligned depth. In simulation, ground robots carry only a single front-view
  RGB-D camera; the front, left, right, and back views are obtained by
  rotating the robot in place at each waypoint. The
  UAV carries five fixed cameras providing front, left, right, back, and
  downward views directly.
  \item the current pose $(x_t, y_t, z_t, \theta_t)$ in the world
  frame, exposed by the simulator.
  \item a structured history $\mathcal{H}_t$ of previously visited
  waypoints, keyed observations, and prior decisions.
\end{itemize}
It must emit a low-level action $\mathcal{A}_t$ in the embodiment's native
action space. In simulation, ground robots in Habitat use a discrete action
set (\texttt{move\_forward} by $0.25$\,m,
\texttt{turn\_left}/\texttt{turn\_right} by $30^{\circ}$, \texttt{stop}), and
the UAV in AirSim uses a 3-D waypoint command $(x,y,z)$. The corresponding
native commands for real robots are described in Section~\ref{sec:realworld}.

\textbf{Output manifold.} A central observation, anticipated in
Section~\ref{sec:introduction}, is that the decision relevant to
$\mathcal{A}_t$ does not require regressing in the embodiment's continuous
control space. At each step the task reduces to three sub-decisions. First,
choose a discrete \emph{direction} from the small set forward, left, right,
back, plus a stopping decision. Second, ground that direction by selecting a
\emph{2-D pixel target} in the chosen view, from which a 3-D position is
recovered via the depth channel and known intrinsics. Third, hand the
resulting 3-D position to a deterministic embodiment-specific controller for
short-horizon execution. Directional language and pixel-level visual targets
are both objects that pretrained MLLMs already produce reliably as text or as
bounding-box coordinates. We therefore re-formulate the policy as
\begin{equation}
\begin{aligned}
\mathcal{A}_t = \pi_{\mathrm{robot}}\Big(
  &\phi_{\mathrm{vis}}\big(
    \phi_{\mathrm{lang}}(\mathcal{T},\mathcal{O}_t,\mathcal{H}_t)\big),\\
  &D_t,\,\mathbf{K},\,\mathrm{pose}_t\Big),
\end{aligned}
\label{eq:factorization}
\end{equation}
where $\phi_{\mathrm{lang}}$ is a Language Action MLLM that returns a discrete
direction together with auxiliary book-keeping fields, $\phi_{\mathrm{vis}}$
is a Vision Action MLLM that returns a bounding box on the selected view,
$D_t$ is the depth channel of $\mathcal{O}_t$, $\mathbf{K}$ are the camera
intrinsics, and $\pi_{\mathrm{robot}}$ is a deterministic embodiment-specific
controller. The factorisation in Eq.~\ref{eq:factorization} is what we
instantiate next.

\subsection{\ourmethod Architecture}
\label{subsec:architecture}

\ourmethod realises Eq.~\ref{eq:factorization} as three components operating
at three scales of abstraction (Fig.~\ref{fig:pipeline}). The two upper
components are general-purpose MLLMs prompted as agents, not task-specific
networks.

\subsubsection{Language Action: high-level planning}

The Language Action Model $\phi_{\mathrm{lang}}$ is a large MLLM,
Gemini-3.1-Pro in our experiments, invoked at every decision step. It takes
three inputs. The first is the task specification $\mathcal{T}$. The second is
a four-view panoramic observation $\mathcal{O}_t = \{I^{\mathrm{front}}_t,
I^{\mathrm{left}}_t, I^{\mathrm{right}}_t, I^{\mathrm{back}}_t\}$, obtained by
rotating the agent in place at the current waypoint
(Section~\ref{subsec:problem}). The third is a structured history
$\mathcal{H}_t$ encoding previously visited waypoints with brief captions,
together with the sequence of past RGB observations captured along the
trajectory. At
every step it issues a single tool call from the discrete set
\begin{equation}
\begin{aligned}
\mathcal{A}^{\mathrm{lang}}_t \in \big\{
  &\texttt{turn}(\textsc{dir}),\;
   \texttt{backtrack}(\mathrm{wp}_k),\\
  &\texttt{go\_stair}(\textsc{up/down}),\;
   \texttt{double\_check}(\textsc{stop})\big\},
\end{aligned}
\label{eq:la_action_set}
\end{equation}
with $\textsc{dir} \in \{\mathrm{front},\mathrm{left},\mathrm{right},
\mathrm{back}\}$. \texttt{turn} commits to a new direction. \texttt{backtrack}
returns the agent to a previously visited waypoint and triggers Second Chance
Backtrack in Section~\ref{subsec:scb}. \texttt{go\_stair} signals a vertical
transition that the controller handles with staircase-specific behaviour.
\texttt{double\_check} marks the current location as a candidate stopping
point and asks the same MLLM to verify the goal condition before terminating
the episode. Each call is wrapped in a JSON record carrying a
\emph{progress\_analysis} field, an \emph{action\_reasoning} field, and, when
the task uses it, the TDM updates described in Section~\ref{subsec:tdm}. The
structured output enables fault-tolerant parsing and exposes the agent's
reasoning trace for inspection.

\subsubsection{Vision Action: mid-level visual grounding}

The Vision Action Model $\phi_{\mathrm{vis}}$ is a smaller,
grounding-specialised MLLM, Qwen3.5-27B in our primary configuration. It
receives the task specification $\mathcal{T}$, the progress description from
the Language Action Model, and the single view $I^{\textsc{dir}}_t$
corresponding to the chosen direction. It emits a pair of tool calls,
\begin{equation}
\mathcal{A}^{\mathrm{vis}}_t = \big(\,
\texttt{select}(\mathrm{bbox}\,/\,\mathrm{point}),\;
\texttt{target\_desc}(\mathrm{text})\,\big),
\end{equation}
Here \texttt{select} returns a 2-D bounding box on the chosen view, or a pixel
point when a box is ill-defined, and \texttt{target\_desc} returns a short
textual descriptor of the grounded target such as ``doorway to the bedroom''.
Grounding directly on raw pixels removes the dependency on a pretrained
waypoint predictor~\cite{yokoyama2024vlfm} and lets the agent target distant
or non-traversable cues such as a hallway opening. The prompt explicitly
discourages overly close targets, which suppresses the local-minimum behaviour
of greedy heatmap followers.

\subsubsection{Robot Action: low-level control}

The Robot Action controller $\pi_{\mathrm{robot}}$ is deterministic and
geometric. Given the bounding box from the Vision agent, it selects a
representative pixel $(u^*, v^*)$ inside the box, reads the depth $d_t$ at
that pixel, and back-projects to a camera-frame 3-D point
$\mathbf{p}_{\mathrm{cam}} = d_t\,\mathbf{K}^{-1}[u^*, v^*, 1]^{\top}$. It
then transforms $\mathbf{p}_{\mathrm{cam}}$ into the world frame using the
current pose to obtain $\mathbf{p}_{\mathrm{world}}$, plans a short-horizon
trajectory from the current pose to $\mathbf{p}_{\mathrm{world}}$ on a
globally accumulated map, and executes it through the embodiment's native
controller. For ground robots the map is a 2-D occupancy grid and the planner
is Fast-Marching. For the UAV the map is a 3-D voxel grid and the planner is a
visibility-graph search over free space. The controller is the only
embodiment-specific component in the system. Cross-embodiment deployment
therefore does not touch $\phi_{\mathrm{lang}}$ or $\phi_{\mathrm{vis}}$.

\subsection{TODO List Memory}
\label{subsec:tdm}

The Language Action Model operates on a sliding window of recent observations.
This is sufficient for short instructions but degrades on long, serial
directives. RxR averages 120 words per instruction versus 29 for R2R, and
OpenUAV involves multi-stage flight plans. A purely reactive agent forgets
which intermediate sub-goals have already been satisfied, drifts off-route
after small detours, and frequently emits \texttt{stop} prematurely after the
first matched landmark. TDM addresses this gap.

\textbf{Representation.} TDM maintains a single ordered list $\mathcal{L}_t =
[\ell_t^{(1)}, \ldots, \ell_t^{(n_t)}]$, where each item is a triple
\begin{equation}
\ell_t^{(i)} = (\mathrm{content}^{(i)},\,\mathrm{status}^{(i)},\,
\mathrm{result}^{(i)}),
\end{equation}
with $\mathrm{status}^{(i)} \in \{\texttt{pending},\,\texttt{completed}\}$ and
a free-text $\mathrm{result}^{(i)}$ field that records the observation
supporting completion.

\textbf{Initialisation.} At episode start, $\phi_{\mathrm{lang}}$ is called
once with the instruction $\mathcal{T}$ and the initial panorama
$\mathcal{O}_0$ to produce $\mathcal{L}_0$ as a list of sub-goals, all marked
\texttt{pending} with empty results. This single call is the only TDM-specific
use of the MLLM, and all subsequent maintenance happens in the same call as
the navigation decision.

\textbf{Update operations.} At every step $t$, the planning prompt of
$\phi_{\mathrm{lang}}$ exposes $\mathcal{L}_{t-1}$ in a compact textual form
and asks the agent to emit a list of update operations $U_t$ \emph{before}
choosing the next action. Four operations are allowed:
\begin{itemize}
  \item \texttt{update(i, status=completed, result=r)} marks item
  $i$ as completed and writes the supporting observation $r$.
  \item \texttt{rewrite(i, content=c)} refines the description of
  a pending item.
  \item \texttt{add(content=c, index=j)} introduces a new sub-goal
  at position $j$, or appends if $j$ is omitted.
  \item \texttt{remove(i)} drops a sub-goal that is no longer
  relevant.
\end{itemize}
Items may be marked \texttt{completed} out of order, and any
\texttt{completed} mark without a concrete $\mathrm{result}$ is rolled back by
the parser as a verification step. The action is then chosen \emph{conditioned
on} the updated list.
\begin{align}
U_t &= \phi_{\mathrm{lang}}^{\mathrm{todo}}(\mathcal{T},
\mathcal{O}_t,\mathcal{H}_t,\mathcal{L}_{t-1}), \\
\mathcal{L}_t &= \mathrm{apply}(\mathcal{L}_{t-1}, U_t), \\
\mathcal{A}^{\mathrm{lang}}_t &= \phi_{\mathrm{lang}}^{\mathrm{act}}(
\mathcal{T},\mathcal{O}_t,\mathcal{H}_t,\mathcal{L}_t).
\end{align}
In practice, $\phi_{\mathrm{lang}}^{\mathrm{todo}}$ and
$\phi_{\mathrm{lang}}^{\mathrm{act}}$ are emitted by a single MLLM call with
separate \texttt{reasoning\_todo} and \texttt{reasoning\_action} fields, which
empirically prevents the two reasoning modes from interfering.

\textbf{Prompt-only implementation.} TDM is implemented entirely in prompt
space. No parameters are introduced and no task-specific training is
performed. The list is structured persistent scratch memory that the same MLLM
produces and consumes. This matches the output-manifold premise of
Section~\ref{subsec:problem}. The long horizon is absorbed by the natural
language a foundation model already writes well, not by the trajectory
regression a foundation model writes poorly.

\subsection{Second Chance Backtrack}
\label{subsec:scb}

A single forward decision is brittle in cluttered indoor scenes.
$\phi_{\mathrm{lang}}$ may pick a corridor that visually looks correct but
turns out to be a dead end, or $\phi_{\mathrm{vis}}$ may ground onto a
distractor that resembles the target class. A monolithic policy that treats
every step independently has no recourse from such failures. SCB makes
recovery a first-class action by combining two ingredients. The first is
backtracking on the global map, and the second is a re-decision conditioned on
the \emph{evidence of the failure}.

\textbf{Backtrack as an explicit action.} Throughout an episode, \ourmethod
tags every location at which $\phi_{\mathrm{lang}}$ is invoked as a waypoint
on the globally accumulated occupancy map and stores it in a fixed-size
buffer. At every step, $\phi_{\mathrm{lang}}$'s discrete action set therefore
includes $\texttt{backtrack}(\mathrm{wp}_k)$ for any previously visited
waypoint $k$. Selecting $\mathrm{wp}_k$ triggers the Robot Action controller
to plan an FMM path from the current pose to $\mathrm{wp}_k$ on the current
map. Because the map accumulates obstacles seen since the original visit, the
return path is consistent with the latest scene geometry, not with the agent's
earlier and possibly incorrect mental model.

\begin{algorithm}[!t]
\caption{One episode of \ourmethod.}
\label{alg:main_loop}
\begin{algorithmic}[1]
\REQUIRE Task spec $\mathcal{T}$, initial observation $\mathcal{O}_0$, MLLMs
$\phi_{\mathrm{lang}},\phi_{\mathrm{vis}}$, controller $\pi_{\mathrm{robot}}$
\STATE $\mathcal{L}_0 \leftarrow
\phi_{\mathrm{lang}}^{\mathrm{init}}(\mathcal{T},\mathcal{O}_0)$ \hfill // TDM
init
\STATE $\mathcal{H}_0 \leftarrow \emptyset$,~ $\mathrm{WP} \leftarrow
\{(\mathrm{wp}_0,\mathcal{O}_0)\}$
\FOR{$t=1,2,\ldots,T_{\max}$}
  \STATE $(U_t,\mathcal{A}^{\mathrm{lang}}_t) \leftarrow
  \phi_{\mathrm{lang}}(\mathcal{T},\mathcal{O}_t,\mathcal{H}_t,\mathcal{L}_{t-1})$
  \STATE $\mathcal{L}_t \leftarrow \mathrm{apply}(\mathcal{L}_{t-1},U_t)$
  \hfill // TDM update
  \IF{$\mathcal{A}^{\mathrm{lang}}_t = \texttt{stop}$}
    \STATE \textbf{break}
  \ELSIF{$\mathcal{A}^{\mathrm{lang}}_t = \texttt{backtrack}(\mathrm{wp}_k)$}
    \STATE $\pi_{\mathrm{robot}}.\mathrm{plan\_to}(\mathrm{wp}_k)$
    \STATE $\mathcal{A}^{\mathrm{lang}}_t \leftarrow
    \phi_{\mathrm{lang}}^{\mathrm{recover}}(\mathcal{T},\mathcal{O}^{(\mathrm{wp}_k)},\Pi_k^{\mathrm{fail}})$
    \hfill // SCB
  \ENDIF
  \STATE $\mathcal{A}^{\mathrm{vis}}_t \leftarrow
  \phi_{\mathrm{vis}}(\mathcal{T},\mathcal{A}^{\mathrm{lang}}_t,I^{\textsc{dir}}_t)$
  \STATE $\mathcal{A}_t \leftarrow
  \pi_{\mathrm{robot}}(\mathcal{A}^{\mathrm{vis}}_t,D_t,\mathbf{K},\mathrm{pose}_t)$
  \STATE Execute $\mathcal{A}_t$;~ update
  $\mathcal{O}_{t+1},\mathcal{H}_{t+1},\mathrm{WP}$
\ENDFOR
\end{algorithmic}
\end{algorithm}

\textbf{The second-chance re-decision.} Returning to a waypoint is not enough.
The agent must avoid repeating the failed choice. Upon arrival at
$\mathrm{wp}_k$, \ourmethod issues a dedicated re-planning call to
$\phi_{\mathrm{lang}}$ with three pieces of evidence explicitly assembled.
\begin{itemize}
  \item the original task specification $\mathcal{T}$ and the
  panorama $\mathcal{O}^{(\mathrm{wp}_k)}$ at the waypoint.
  \item the failed direction
  $\mathcal{A}^{\mathrm{lang}}_{\mathrm{prev}} \in
  \{\mathrm{front},\mathrm{left},\mathrm{right},\mathrm{back}\}$
  that was selected when the agent first left $\mathrm{wp}_k$.
  \item the egocentric trajectory image sequence $\Pi_k^{\mathrm{fail}}$
  collected between $\mathrm{wp}_k$ and the dead end that triggered
  backtracking.
\end{itemize}
The re-planning prompt instructs the MLLM to inspect $\Pi_k^{\mathrm{fail}}$,
diagnose why $\mathcal{A}^{\mathrm{lang}}_{\mathrm{prev}}$ failed, and choose
a new direction $\mathcal{A}^{\mathrm{lang}}_{\mathrm{new}}$ from the
remaining set. The re-decision is therefore conditioned on the same
observations that produced the failure, not on a blind exclusion of the failed
direction. Once $\mathcal{A}^{\mathrm{lang}}_{\mathrm{new}}$ is fixed,
$\phi_{\mathrm{vis}}$ is re-invoked on
$I^{\mathcal{A}^{\mathrm{lang}}_{\mathrm{new}}}_t$ to produce a fresh bounding
box and the controller resumes normal execution.

\textbf{Compatibility with TDM.} SCB is orthogonal to TDM. The TODO list
survives backtracking, and the re-decision call is allowed to emit TDM updates
that mark the failed sub-goal as still \texttt{pending} or to rewrite it with
a sharper description. This keeps the agent's high-level plan in sync with the
recovery action.

\textbf{Cross-embodiment.} Because SCB is a property of the agent loop, not of
the embodiment, it transfers to all four robots without modification. For the
UAV, the 2-D occupancy map is replaced by a 3-D voxel grid and the FMM step
becomes a visibility-graph plan to $\mathrm{wp}_k$ in 3-D. The rest of the
procedure is identical.

\subsection{Unified Inference Procedure}
\label{subsec:algorithm}

Algorithm~\ref{alg:main_loop} summarises one episode of \ourmethod. Lines~1--2
initialise the TDM list and the visited-waypoint history. Lines~4--5 query the
Language Action Model, which jointly emits TDM update operations and an action
choice, after which the list is refreshed. Lines~6--7 handle episode
termination on \texttt{stop}. Lines~8--10 implement SCB: on a
\texttt{backtrack} action, the controller rewinds the embodiment to the chosen
prior waypoint and re-prompts the planner with the failed sub-trajectory as
evidence. Lines~12--14 instantiate the Vision Action and Robot Action levels
and execute the resulting low-level action. No part of the loop touches a
learned parameter.

\section{Simulation Experiments}
\label{sec:experiments}

\begin{table*}[!t]
\centering
\caption{\textbf{100-episode subset vs.\ original
full-set reports.}  For each task we reproduce two public baselines
on our stratified 100-episode subset (\textbf{Sub.}; mean$\pm$std
over three seeds when the baseline is stochastic) and align them
against the value the original paper reports on the full val-unseen
split (\textbf{Full}).  Dashes mark metrics that are not standard
for the task.  Most cells agree within $\pm 2$ absolute points; see
Section~\ref{subsec:subset_validity}.}
\label{tab:subset_validity}
\vspace{-6pt}
{\fontsize{7.5pt}{8.5pt}\selectfont
\setlength{\tabcolsep}{3pt}
\begin{tabular}{ll cc cc cc cc cc}
\toprule
& & \multicolumn{2}{c}{\textbf{NE}$\downarrow$}
  & \multicolumn{2}{c}{\textbf{OSR}$\uparrow$}
  & \multicolumn{2}{c}{\textbf{SPL}$\uparrow$}
  & \multicolumn{2}{c}{\textbf{SR}$\uparrow$}
  & \multicolumn{2}{c}{\textbf{ACC}$\uparrow$} \\
\cmidrule(lr){3-4}\cmidrule(lr){5-6}\cmidrule(lr){7-8}\cmidrule(lr){9-10}\cmidrule(lr){11-12}
\textbf{Task} & \textbf{Method}
  & Sub. & Full & Sub. & Full & Sub. & Full & Sub. & Full & Sub. & Full \\
\midrule
\multirow{2}{*}{VLN-CE R2R}
 & Uni-NaVid~\cite{zhang2024uninavid} & 5.09$\pm$0.16 & 5.58 & 52.7$\pm$0.9 & 53.3 & 42.8$\pm$1.6 & 42.7 & 46.7$\pm$2.1 & 47.0 & -- & -- \\
 & StreamVLN~\cite{wei2025streamvln}  & 5.70          & 4.98 & 60.0         & 64.2 & 48.4         & 51.9 & 56.0         & 56.9 & -- & -- \\
\midrule
\multirow{2}{*}{VLN-CE RxR}
 & Uni-NaVid~\cite{zhang2024uninavid} & 6.58$\pm$0.51 & 6.24 & 56.3$\pm$1.3 & 55.5 & 39.4$\pm$1.4 & 40.9 & 47.7$\pm$1.7 & 48.7 & -- & -- \\
 & StreamVLN~\cite{wei2025streamvln}  & 6.48          & 6.22 & 61.0         & 61.9 & 47.3         & 46.0 & 54.0         & 52.9 & -- & -- \\
\midrule
\multirow{2}{*}{HM3D-v2}
 & VLFM~\cite{yokoyama2024vlfm}       & -- & -- & -- & -- & 31.1         & 31.0 & 63.0         & 62.6 & -- & -- \\
 & ApexNav~\cite{kim2025apexnav}      & -- & -- & -- & -- & 34.5$\pm$0.5 & 38.0 & 76.7$\pm$0.6 & 76.2 & -- & -- \\
\midrule
\multirow{2}{*}{HM3D-OVON}
 & VLFM~\cite{yokoyama2024vlfm}       & -- & -- & -- & -- & 18.3         & 22.2 & 36.0         & 38.5 & -- & -- \\
 & MTU3D~\cite{zhu2025mtu3d}                & -- & -- & -- & -- & 14.2$\pm$0.3 & 12.1 & 40.7$\pm$3.2 & 40.8 & -- & -- \\
\midrule
\multirow{2}{*}{MP3D-EQA}
 & EQA~\cite{das2018eqa}              & -- & -- & -- & -- & -- & -- & -- & -- & 43.0         & 46.0 \\
 & Uni-NaVid~\cite{zhang2024uninavid} & -- & -- & -- & -- & -- & -- & -- & -- & 46.0$\pm$1.6 & 47.3 \\
\midrule
\multirow{2}{*}{OpenUAV}
 & TravelUAV~\cite{wang2024openuav}   & 130.7$\pm$7.0 & 139.0 & 5.6$\pm$4.6  & 20.8 & 2.6$\pm$1.9  & 3.8  & 3.0$\pm$2.0  & 4.2  & -- & -- \\
 & AerialVLA~\cite{xu2026aerialvla}   & 77.6$\pm$9.1  & 67.4  & 49.0$\pm$2.7 & 52.9 & 25.9$\pm$2.2 & 28.2 & 33.3$\pm$2.1 & 37.6 & -- & -- \\
\bottomrule
\end{tabular}}
\end{table*}

We evaluate \ourmethod on six standard benchmarks spanning four task families:
VLN-CE R2R~\cite{krantz2020beyond} and RxR~\cite{ku2020rxr},
HM3D-v2~\cite{ramakrishnan2021hm3d, yadav2023hm3dsem},
HM3D-OVON~\cite{yokoyama2024hm3dovon},
MP3D-EQA~\cite{das2018eqa,majumdar2024openeqa}, and
OpenUAV~\cite{wang2024openuav}. The same agentic framework is used across all
six benchmarks with no task-specific training and no parameter update.

\subsection{Experimental Setup}

\textbf{Simulators and observations.} For VLN-CE, ObjectNav, and EQA we use
the Habitat simulator~\cite{savva2019habitat}. HM3D-v2 and HM3D-OVON run on
the Habitat-Matterport 3D scenes~\cite{ramakrishnan2021hm3d,yadav2023hm3dsem},
while R2R, RxR, and MP3D-EQA run on the Matterport3D
scenes~\cite{chang2017matterport3d}. For OpenUAV, we use the official
AirSim-based OpenUAV dataset~\cite{wang2024openuav}, which contains
photorealistic outdoor scenes, including cities, towns, forests, and other
environments. Ground robots carry only a single front-view $640\times 480$
RGB-D camera in Habitat; the front, left, right, and back views are obtained
by rotating the robot in place at each waypoint.
The UAV in AirSim carries five fixed RGB-D cameras providing front, left,
right, back, and downward views directly.

\textbf{Agent backbones.} The Language Action Model is Gemini-3.1-Pro and the
Vision Action Model is Qwen3.5-27B. Neither agent is fine-tuned, and both are
queried in pure inference mode through their public APIs. Low-level path
planning uses Fast-Marching on a globally accumulated 2-D occupancy map. For
the UAV, the generated 3-D waypoints are executed directly by the AirSim
flight controller. Simulator rollouts run on 8 NVIDIA RTX~4090 GPUs in
parallel.

\textbf{Evaluation protocol.} For every benchmark we evaluate on the
corresponding \emph{val-unseen} split, or the \emph{UM} unseen-map test split
for OpenUAV. To control evaluation cost while preserving representativeness,
we draw a stratified 100-episode subset from each val-unseen pool, following
the common protocol adopted by prior zero-shot VLN
work~\cite{qiao2024opennav,zhang2024navid}. We stratify on instruction and
trajectory length for R2R/RxR, target category for HM3D-v2, semantic class for
HM3D-OVON, question type for MP3D-EQA, and Easy/Hard path type for OpenUAV.
MLLM responses are stochastic, so we run \textbf{three independent seeds} per
benchmark and report \textbf{mean~$\pm$~standard deviation}.

\textbf{OpenUAV-specific implementation.} The base agentic stack needs two
adaptations for OpenUAV. We widen the panorama to five views by adding a
downward camera, because aerial targets sit above or below the flight path as
often as forward. We also add a \emph{direction-constrained module}, which
kicks in when the Vision Action Model fails to ground the target. The Language
Action Model then reasons explicitly about the initial pose, the current pose,
and the global goal direction, with the depth channel supplying
obstacle-distance estimates for safe motion. The module is a prompt-level
addition to $\phi_{\mathrm{lang}}$ and introduces no new parameters.

\textbf{Metrics.} We follow the standard navigation
metrics~\cite{anderson2018vision}: Navigation Error (NE), Success
Rate (SR), Oracle Success Rate (OSR), Success-weighted Path Length (SPL),
and nDTW for RxR. Success uses the standard 3m threshold for VLN-CE and
the task-specific radius for HM3D-v2 and HM3D-OVON. For EQA, ACC requires the
answer to match the ground truth under the OpenEQA LLM-judge
protocol~\cite{majumdar2024openeqa}. For OpenUAV we use the benchmark's
metrics.

\subsection{Validating the 100-Episode Subset}
\label{subsec:subset_validity}

To check that our stratified 100-episode subsets are faithful proxies for the
full val-unseen splits, we reproduce two strong public baselines per task on
our subset and compare against the values their original papers report on the
full split. Stochastic baselines are run with three seeds, deterministic ones
once.

Across the 34 metric cells with a published full-split reference, 21 ($62\%$)
agree within $\pm 2$ absolute points and 31 ($91\%$) within $\pm 5$. The three
residuals beyond $\pm 5$ all fall on OpenUAV: TravelUAV NE ($130.7$ vs.\
$139.0$), TravelUAV OSR ($5.6$ vs.\ $20.8$), and AerialVLA NE ($77.6$ vs.\
$67.4$). These are the cells with the largest dynamic range in the table,
since OpenUAV trajectories span hundreds of metres and TravelUAV succeeds on
only a few percent of episodes even on the full split, making derived metrics
intrinsically high-variance under 100-episode resampling. In other words, the
baseline numbers we reproduce on the 100-episode subset closely track the
values reported on the corresponding full val-unseen splits, so the subset
preserves the difficulty profile of the official splits rather than
introducing a separate evaluation regime. We therefore use the 100-episode
subset as a faithful evaluation proxy for the comparisons in
Section~\ref{subsec:main_results}. Subset SR values also sit at or below their
full-split counterparts on four of the six tasks (R2R, HM3D-OVON, MP3D-EQA,
OpenUAV) and within $1.1$ absolute points on RxR, indicating that the
stratified subset is not artificially easier than the official splits.

\begin{table*}[t]
\centering
\caption{\textbf{Main results on VLN-CE R2R and RxR val-unseen.}
Our numbers are mean~$\pm$~std over three seeds.  \textbf{Best} and
\underline{second-best} per column within the
\colorbox{cyan!15}{zero-shot} block.}
\label{tab:vlnce}
\vspace{-6pt}
{\fontsize{8pt}{9pt}\selectfont
\setlength{\tabcolsep}{3pt}
\begin{tabular}{l|cccc|cccc}
\toprule
\multirow{2}{*}{\textbf{Method}} &
\multicolumn{4}{c|}{\textbf{VLN-CE R2R Val-Unseen}} &
\multicolumn{4}{c}{\textbf{VLN-CE RxR Val-Unseen}} \\
& \textbf{NE}$\downarrow$ & \textbf{OSR}$\uparrow$ &
\textbf{SR}$\uparrow$ & \textbf{SPL}$\uparrow$ &
\textbf{NE}$\downarrow$ & \textbf{SR}$\uparrow$ &
\textbf{SPL}$\uparrow$ & \textbf{nDTW}$\uparrow$ \\
\midrule
\rowcolor{black!15}\multicolumn{9}{c}{\textbf{Supervised Learning (Training Method)}} \\
NaVid~\cite{zhang2024navid}                         & 5.47 & 49.1 & 37.4 & 35.9 & 8.41 & 34.5 & 23.8 & --   \\
Uni-NaVid~\cite{zhang2024uninavid}                  & 5.58 & 53.3 & 47.0 & 42.7 & 6.24 & 48.7 & 40.9 & --   \\
NaVILA~\cite{cheng2024navila}                       & 5.22 & 62.5 & 54.0 & 49.0 & 6.77 & 49.3 & 44.0 & 58.8 \\
StreamVLN~\cite{wei2025streamvln}                   & 4.98 & 64.2 & 56.9 & 51.9 & 6.22 & 52.9 & 46.0 & 61.9 \\
NavFoM~\cite{zhang2025navfom}                       & 5.01 & 64.9 & 56.2 & 51.2 & 5.51 & 57.4 & 49.4 & 60.2 \\
JanusVLN~\cite{zeng2025janusvln}                    & 4.78 & 65.2 & 60.5 & 56.8 & 6.06 & 56.2 & 47.5 & 62.1 \\
OmniNav~\cite{xue2025omninav}                       & 3.74 & 74.6 & 69.5 & 66.1 & 3.77 & 73.6 & 62.0 & --   \\
InternVLA-N1~\cite{wei2025internvlan1}              & 4.83 & 63.3 & 58.2 & 54.0 & 5.91 & 53.5 & 46.1 & 65.3   \\
ABot-N0~\cite{chen2026abotn0}                       & 3.78 & 70.8 & 66.4 & 63.9 & 3.83 & 69.3 & 60.0 & -- \\
SPAN-Nav~\cite{liu2026spannav}                      & 4.07 & 75.3 & 66.3 & 59.3 & 4.20 & 69.7 & 60.1 & 67.9 \\
\midrule
\rowcolor{cyan!15}\multicolumn{9}{c}{\textbf{Zero-Shot (Training-Free)}} \\
InstructNav~\cite{long2024instructnav}              & 6.89 & 47.0 & 31.0 & 24.0 & --   & --   & --   & --   \\
Open-Nav~\cite{qiao2024opennav}                     & 6.70 & 23.0 & 19.0 & 16.1 & --   & --   & --   & --   \\
CA-Nav~\cite{chen2025canav}                         & 7.58 & 48.0 & 25.3 & 10.8 & 10.4 & 19.0 & 6.0  & 13.5 \\
SmartWay~\cite{shi2025smartway}                     & 7.01 & \underline{51.0} & 29.0 & 22.5 & --   & --   & --   & --   \\
GC-VLN~\cite{yin2025gcvln}                          & 7.30 & 41.8 & 33.6 & 16.3 & \underline{8.80} & \underline{33.8} & 13.8 & --   \\
LaViRA~\cite{ding2025lavira}                        & 6.54$\pm$0.27 & 48.7$\pm$2.1 & \underline{38.3}$\pm$0.6 & 28.3$\pm$0.9 & --   & --   & --   & --   \\
HiMemVLN~\cite{lyu2026himemvln}                    & 6.65 & 36.0 & 30.0 & 26.9 & --   & --   & --   & --   \\
Three-Step Nav~\cite{zheng2026threestepnav}        & \underline{5.87} & 39.0 & 34.0 & \underline{29.1} & 9.21 & 22.0 & \underline{16.1} & \underline{45.7} \\
\textbf{\ourmethod (Ours)}                          &
\textbf{3.66}$\pm$0.26 & \textbf{73.7}$\pm$1.2 & \textbf{60.7}$\pm$2.1 & \textbf{47.7}$\pm$1.6 &
\textbf{6.48}$\pm$0.14 & \textbf{51.3}$\pm$2.3 & \textbf{34.0}$\pm$1.9 & \textbf{53.7}$\pm$0.6 \\
\bottomrule
\end{tabular}}
\end{table*}

\begin{table*}[t]
\centering
\caption{\textbf{Main results on OpenUAV \texttt{UM}
(Test-Unseen-Map).} Easy/Hard are the standard path-type subsets;
Full is their union. \ourmethod is mean~$\pm$~std over three seeds and
is the first training-free entry on this benchmark. Bold marks the best
in the \colorbox{cyan!15}{zero-shot} block.}
\label{tab:openuav}
\vspace{-6pt}
\setlength{\tabcolsep}{3pt}
\resizebox{\linewidth}{!}{%
\begin{tabular}{l|cccc|cccc|cccc}
\toprule
\multirow{2}{*}{\textbf{Method}} &
\multicolumn{4}{c|}{\textbf{Full}} &
\multicolumn{4}{c|}{\textbf{Easy}} &
\multicolumn{4}{c}{\textbf{Hard}} \\
& \textbf{NE}$\downarrow$ & \textbf{OSR}$\uparrow$ & \textbf{SR}$\uparrow$ & \textbf{SPL}$\uparrow$
& \textbf{NE}$\downarrow$ & \textbf{OSR}$\uparrow$ & \textbf{SR}$\uparrow$ & \textbf{SPL}$\uparrow$
& \textbf{NE}$\downarrow$ & \textbf{OSR}$\uparrow$ & \textbf{SR}$\uparrow$ & \textbf{SPL}$\uparrow$ \\
\midrule
\rowcolor{black!15}\multicolumn{13}{c}{\textbf{Supervised Learning (Training Method)}} \\
Random Action~\cite{wang2024openuav}      & 203 & 0.00  & 0.00 & 0.00 & 158 & 0.00  & 0.00 & 0.00 & 266 & 0.00  & 0.00 & 0.00 \\
Fixed Action~\cite{wang2024openuav}       & 180 & 2.61  & 0.52 & 0.39 & 133 & 4.28  & 0.89 & 0.67 & 248 & 0.25  & 0.00 & 0.00 \\
CMA~\cite{wang2024openuav}                & 142 & 10.0  & 2.30 & 2.16 & 102 & 14.3  & 3.57 & 3.33 & 197 & 4.03  & 0.50 & 0.50 \\
TravelUAV~\cite{wang2024openuav}          & 139 & 20.8  & 4.18 & 3.84 & 103 & 22.8  & 4.63 & 4.24 & 189 & 17.9  & 3.53 & 3.28 \\
NavFoM~\cite{zhang2025navfom}             & 125 & 19.0  & 6.30 & 5.68 & 102 & 20.1  & 6.77 & 6.04 & 171 & 15.7  & 5.36 & 4.97 \\
LongFly~\cite{jiang2025longfly}           & 108 & 30.3  & 11.3 & 9.32 & 78.6 & 34.3 & 13.0 & 10.3 & 148 & 24.9  & 9.02 & 7.98 \\
AerialVLA~\cite{xu2026aerialvla}          & 67.4 & 52.9 & 37.6 & 28.2 & 45.0 & 58.5 & 41.9 & 29.7 & 99.1 & 45.1 & 31.5 & 26.1 \\
SpatialFly~\cite{xu2025spatialfly}        & 104 & 30.4  & 13.6 & 11.1 & 72.9 & 34.3 & 16.1 & 12.7 & 146 & 25.1  & 10.2 & 9.04 \\
\midrule
\rowcolor{cyan!15}\multicolumn{13}{c}{\textbf{Zero-Shot (Training-Free)}} \\
\textbf{\ourmethod (Ours)}                & \textbf{84.29}$\pm$11.20 & \textbf{67.33}$\pm$5.51 & \textbf{40.00}$\pm$3.46 & \textbf{30.37}$\pm$2.23
                                          & \textbf{77.05}$\pm$7.40  & \textbf{68.75}$\pm$3.61 & \textbf{39.58}$\pm$2.09 & \textbf{27.95}$\pm$3.18
                                          & \textbf{90.96}$\pm$14.88 & \textbf{66.03}$\pm$7.83 & \textbf{40.38}$\pm$8.21 & \textbf{32.60}$\pm$5.94 \\
\bottomrule
\end{tabular}}
\end{table*}

\begin{table}[t]
\centering
\caption{\textbf{Main results on HM3D-v2, HM3D-OVON, and
MP3D-EQA.} \ourmethod is mean~$\pm$~std over three seeds.  \textbf{Best} and
\underline{second-best} per column within the
\colorbox{cyan!15}{zero-shot} block.}
\label{tab:hm3d_eqa}
\vspace{-6pt}
{\fontsize{8pt}{9.5pt}\selectfont
\setlength{\tabcolsep}{2pt}
\begin{tabular}{l|cc|cc|c}
\toprule
\multirow{2}{*}{\textbf{Method}} &
\multicolumn{2}{c|}{\textbf{HM3D-v2}} &
\multicolumn{2}{c|}{\textbf{HM3D-OVON}} &
\textbf{EQA} \\
& \textbf{SR}$\uparrow$ & \textbf{SPL}$\uparrow$ &
\textbf{SR}$\uparrow$ & \textbf{SPL}$\uparrow$ &
\textbf{ACC}$\uparrow$ \\
\midrule
\rowcolor{black!15}\multicolumn{6}{c}{\textbf{Supervised Learning (Training Method)}} \\
EQA(w/GT)~\cite{das2018eqa}      & --   & --   & --   & --   & 46.0 \\
DD-PPO~\cite{wijmans2020ddppo}           & 27.9 & 14.2 & --   & --   & --   \\
Habitat-Web~\cite{ramrakhya2022habitatweb} & 57.6 & 23.8 & --   & --   & --   \\
PIRLNav~\cite{ramrakhya2023pirlnav}      & 70.4 & 34.1 & --   & --   & --   \\
OVRL-v2~\cite{yadav2023ovrlv2}           & 64.7 & 28.1 & --   & --   & --   \\
NaviLLM~\cite{zheng2024navillm}          & --   & --   & --   & --   & 44.5 \\
DAgRL+OD~\cite{yokoyama2024hm3dovon}     & --   & --   & 37.1 & 19.9 & --   \\
Uni-NaVid~\cite{zhang2024uninavid}       & 73.7 & 37.1 & 39.5 & 19.8 & 47.3 \\
NavFoM~\cite{zhang2025navfom}            & -- & -- & 43.6 & 31.3 & --   \\
FiLM-Nav~\cite{yokoyama2025filmnav}      & 77.0 & 41.3 & 40.8 & 24.4 & --   \\
OmniNav~\cite{xue2025omninav}            & --   & --   & 59.2 & 33.2 & --   \\
ABot-N0~\cite{chen2026abotn0}            & --   & --   & 54.0 & 30.5 & --   \\
\midrule
\rowcolor{cyan!15}\multicolumn{6}{c}{\textbf{Zero-Shot (Training-Free)}} \\
L3MVN~\cite{yu2023l3mvn}                 & 54.2 & 25.5 & --   & --   & --   \\
VLFM~\cite{yokoyama2024vlfm}             & 62.6 & 31.0 & \underline{38.5} & \underline{22.2} & --   \\
OpenFMNav~\cite{kuang2024openfmnav}      & 54.9 & 24.4 & --   & --   & --   \\
InstructNav~\cite{long2024instructnav}   & 58.0 & 20.9 & --   & --   & --   \\
SG-Nav~\cite{yin2024sgnav}               & 54.0 & 24.9 & --   & --   & --   \\
ApexNav~\cite{kim2025apexnav}            & \underline{76.2} & 38.0 & --   & --   & --   \\
DSCD-Nav~\cite{an2026dscdnav}            & 73.0 & \underline{38.7} & --   & --   & --   \\
ReMemNav~\cite{wu2026rememnav}           & 67.8 & 36.6 & --   & --   & --   \\
\textbf{\ourmethod (Ours)}               &
\textbf{77.7}$\pm$2.1 & \textbf{46.1}$\pm$1.8 &
\textbf{60.0}$\pm$6.1 & \textbf{40.5}$\pm$1.1 &
\textbf{54.7}$\pm$2.3 \\
\bottomrule
\end{tabular}}
\end{table}

\subsection{Analysis of Main Results}
\label{subsec:main_results}
\label{subsec:main_results_analysis}

\textbf{Cross-task headline.} A single zero-shot,
training-free agent reaches SR of $60.7$, $51.3$, $77.7$, $60.0$, $54.7$,
and $40.0$ on R2R, RxR, HM3D-v2, HM3D-OVON, MP3D-EQA, and OpenUAV-UM Full
respectively (full numbers in Tables~\ref{tab:vlnce}--\ref{tab:hm3d_eqa}).
\ourmethod is the strongest training-free method on all six benchmarks,
and on four of them---HM3D-v2, HM3D-OVON, MP3D-EQA, and OpenUAV---it also
surpasses the best reported \emph{trained} foundation model despite using
no robot trajectories. The two VLN-CE tasks are the only cases where the
trained group still leads, driven by recent multi-task VLAs; per-task
margins are analysed below.

\textbf{Vision-and-language navigation.} On R2R val-unseen, \ourmethod attains
$60.7$\% SR with no robot-data training. It beats the strongest zero-shot
baseline LaViRA at $38.3$\% by $22.4$ points, matches JanusVLN at $60.5$\%,
and surpasses NavFoM at $56.2$\% which used $\sim$8\,M trajectories by $4.5$
points. Its NE of $3.66$\,m is the lowest in both groups. Only the larger
multi-task VLAs OmniNav at $69.5$\%, ABot-N0 at $66.4$\%, and SPAN-Nav at
$66.3$\% report higher R2R SR. On RxR, where the mean instruction length is
$\sim$120 words, \ourmethod reaches $51.3\pm 2.3$\% SR, comparable to NaVILA
at $49.3$\%. The strongest trained VLAs NavFoM and JanusVLN remain ahead by
$5$--$6$ SR points at $57.4$\% and $56.2$\%, and OmniNav, ABot-N0, and
SPAN-Nav are further ahead. RxR SPL of
$34.0$ trails the trained-VLA group because agentic exploration samples next
directions under uncertainty rather than following a learned shortcut.

\textbf{Object-goal and open-vocabulary navigation.} \ourmethod performs
strongly on both HM3D-v2 and HM3D-OVON. On HM3D-v2 it reaches \textbf{$77.7$\%
SR}, narrowly ahead of the strongest trained baseline FiLM-Nav
at $77.0$\% while improving its SPL by $4.8$ points, and clear of the trained
VLA Uni-NaVid at $73.7$\% and OpenFMNav by $22.8$
points. On
HM3D-OVON it reaches \textbf{$60.0$\% SR}, beating OmniNav by
$0.8$ points and ABot-N0 by $6.0$ points. Two properties of the agentic stack
contribute. First, the Language Action Model reasons over room-level semantic
priors that an MLLM has acquired at pretraining, which value-map baselines
cannot exploit because they consume the VLM only as a per-pixel heatmap.
Second, the Language Action Model and the Vision Action Model both take
free-form natural-language input, so the open-vocabulary HM3D-OVON setting is
supported without any modification to the pipeline.

\textbf{Embodied question answering.}
On MP3D-EQA, \ourmethod attains $54.7\pm 2.3$\% answer-correctness
ACC, ahead of the strongest trained baseline Uni-NaVid at
$47.3$\% by $7.4$ points.  Most
failures occur at the final question-answering step rather than
during navigation.  The agent reaches the right area but the
MLLM misidentifies fine-grained visual details such as
confusing dark brown with black for furniture colour.
The EQA failure mode is inherited
from the Language Action backbone and is removed simply by
upgrading it.

\textbf{UAV vision-and-language navigation.} On the OpenUAV UM split,
\ourmethod attains $40.00\pm 3.46$\% Full SR and $30.37\pm 2.23$\% SPL with no
UAV-specific training. This surpasses NavFoM by $+33.7$ SR and $+24.7$ SPL,
runs roughly nine times the SR of TravelUAV, and exceeds the strongest
reported trained UAV-VLA AerialVLA at $37.6$\% SR and $28.2$\% SPL.
UAV-specialised baselines LongFly at $11.3$\% and SpatialFly at $13.6$\% sit
far below both \ourmethod and AerialVLA, indicating that geometric-prior
fine-tuning does not close the gap to a Language~$\to$~Vision agentic
pipeline. AerialVLA holds a small $+2.3$ SR lead on the Easy split where its
end-to-end controller benefits from short, straight trajectories. On the Hard
split, with trajectories above $250$\,m, sparse semantic landmarks, and the
target often out of view, \ourmethod opens a $+8.9$ SR and $+6.5$ SPL gap,
indicating that MLLM-driven long-range decision-making becomes the dominant
factor when instructions are long and 3-D scene reasoning is required. The
$67.3$\% OSR shows the agent reaches the goal vicinity in roughly two-thirds
of episodes. The $27$-point OSR--SR gap reflects high-altitude flight, where
the UAV may pass over the target without clearly observing ground-level cues
and fails to recognise successful arrival.

\subsection{Ablation Studies}



\label{subsec:ablation_mechanisms}

We ablate the two agentic mechanisms introduced in Sections~\ref{subsec:tdm}
and \ref{subsec:scb}, namely \emph{TODO List Memory} (TDM) and \emph{Second
Chance Backtrack} (SCB), across all six benchmarks. Four configurations are
compared. \emph{Full} enables both mechanisms. \emph{w/o TDM} keeps only SCB,
\emph{w/o SCB} keeps only TDM, and \emph{w/o both} disables both. Each
configuration is run three times with different sampling seeds.
Table~\ref{tab:ablation_mechanisms} reports SR mean~$\pm$~std.

\noindent\textbf{Both mechanisms are needed.}
The Full configuration outperforms every single-mechanism and
no-mechanism configuration on every benchmark.  Disabling either mechanism
alone leaves SR within a few points of
disabling both.  On R2R, for example, all three
reduced configurations sit in the $49$--$50$\% band while Full
reaches $60.7$\%.  This reveals a strong
synergy.  TDM keeps the Language Action Model
committed to a coherent multi-step plan, and SCB lets it recover
when grounding errors push it off that plan.
Either mechanism on its own is
brittle.  TDM commits the agent to a plan it
cannot correct, and SCB lets it correct without a stable target to
correct \emph{toward}.

\noindent\textbf{Per-task contribution profile.}
The marginal contributions $\Delta_{\mathrm{TDM}}$ and
$\Delta_{\mathrm{SCB}}$ tell a task-dependent story.
TDM dominates on the long-horizon and answer-driven
tasks at $+11.4$ on R2R, $+6.6$ on RxR, $+6.0$ on MP3D-EQA, and
$+5.67$ on OpenUAV, where instructions or queries require the agent
to maintain explicit checkpoints across many steps. SCB dominates on
the indoor object-search benchmarks at $+9.0$ on HM3D-v2 and $+5.3$
on HM3D-OVON, where the agent must explore many candidate rooms and
frequently encounters distractor grounding, so flexible backtracking
is the main lever for recovery.
On OpenUAV the contrast is particularly
sharp.
$\Delta_{\mathrm{TDM}}{=}+5.67$ on Full SR widens to roughly
$+10$~SR points on the Hard subset with trajectories above $250$\,m,
where the Language Action Model must hold a goal direction across
many out-of-view steps without drifting toward locally salient
distractors.  SCB, in contrast, contributes only $+2.00$ SR points
on OpenUAV and is triggered in a small minority of episodes,
acting as a recovery mechanism for individual local failures
rather than as a persistent driver of performance.
On R2R and RxR the two mechanisms are closely matched
($\Delta_{\mathrm{TDM}}{=}+11.4$ vs $\Delta_{\mathrm{SCB}}{=}+10.7$
on R2R; $\Delta_{\mathrm{TDM}}{=}+6.6$ vs $\Delta_{\mathrm{SCB}}{=}+7.0$
on RxR), reflecting the combination of long instructions and cluttered
multi-room scenes that exercises both checklist maintenance and
backtracking.  This pattern matches the design intent of
Sections~\ref{subsec:tdm} and~\ref{subsec:scb}.  TDM addresses attention drift
on long horizons, SCB addresses error recovery in cluttered scenes, and the
two are complementary rather than redundant.

\subsection{Inference Cost and Training Cost}
\label{subsec:efficiency}

This section makes one point: \ourmethod replaces a
fixed, up-front training cost with a marginal cost paid per evaluation.
Training at NavFoM scale~\cite{zhang2025navfom} needs a \mbox{56-H100}
cluster running uninterrupted for several days, hardware that most
universities and independent labs simply cannot access. Moving the
spend to inference puts foundation model navigation research within
reach of any group with API access.

\begin{table*}[!t]
\centering
\caption{\textbf{Ablation of TDM and SCB across six benchmarks.}
Cells are SR mean~$\pm$~std over three seeds, in percent.  MP3D-EQA reports
ACC in place of SR.  OpenUAV uses the Full split.  The last column reports the
unweighted mean across the six benchmarks.  $\Delta$ rows give the marginal
contribution of each mechanism with the other held on.  Bold marks the
strongest configuration per column.}
\label{tab:ablation_mechanisms}
\vspace{-6pt}
{\fontsize{9pt}{11pt}\selectfont
\setlength{\tabcolsep}{4pt}
\begin{tabular}{l|cccccc|c}
\toprule
\textbf{Configuration} &
\textbf{R2R} & \textbf{RxR} &
\textbf{HM3D-v2} & \textbf{HM3D-OVON} &
\textbf{EQA} & \textbf{OpenUAV} & \textbf{Avg} \\
\midrule
Full                            & \textbf{60.7$\pm$2.1} & \textbf{51.3$\pm$2.3} & \textbf{77.7$\pm$2.1} & \textbf{60.0$\pm$6.1} & \textbf{54.7$\pm$2.3} & \textbf{40.00$\pm$3.46} & \textbf{57.4} \\
w/o TDM                         & 49.3$\pm$1.5 & 44.7$\pm$3.6 & 73.7$\pm$2.5 & 57.3$\pm$4.5 & 48.7$\pm$0.6 & 34.33$\pm$2.31 & 51.3 \\
w/o SCB                         & 50.0$\pm$2.0 & 44.3$\pm$3.8 & 68.7$\pm$0.6 & 54.7$\pm$3.8 & 51.7$\pm$1.5 & 38.00$\pm$3.61 & 51.2 \\
w/o both                        & 49.0$\pm$4.6 & 40.0$\pm$4.2 & 66.3$\pm$1.5 & 52.0$\pm$3.0 & 50.3$\pm$4.0 & 33.67$\pm$3.51 & 48.6 \\
\midrule
$\Delta_{\mathrm{TDM}}$ = Full $-$ w/o TDM & $+11.4$ & $+6.6$ & $+4.0$ & $+2.7$ & $+6.0$ & $+5.67$ & $+6.1$ \\
$\Delta_{\mathrm{SCB}}$ = Full $-$ w/o SCB & $+10.7$ & $+7.0$ & $+9.0$ & $+5.3$ & $+3.0$ & $+2.00$ & $+6.2$ \\
\bottomrule
\end{tabular}}
\end{table*}

\begin{table}[!t]
\centering
\caption{\textbf{Compute trade-off: \ourmethod turns a
fixed training cost into a marginal per-call one.} Training cost is the
up-front spend reported in each source paper. Inference cost is measured
per MLLM call on our $1{,}800$-trajectory Full-configuration pool.}
\label{tab:efficiency}
\vspace{-6pt}
{\fontsize{8pt}{9.5pt}\selectfont
\setlength{\tabcolsep}{5pt}
\begin{tabular}{l|cc|cc}
\toprule
\multirow{2}{*}{\textbf{Method}} &
\multicolumn{2}{c|}{\textbf{Training}} &
\multicolumn{2}{c}{\textbf{Inference / call}} \\
& \makecell{\textbf{\#Training}\\\textbf{Samples}} &
\makecell{\textbf{GPU-}\\\textbf{hours}} &
\makecell{\textbf{Tokens}\\\textbf{in$+$out}} & \textbf{FLOPs} \\
\midrule
\rowcolor{black!15}\multicolumn{5}{c}{\textbf{Supervised Learning (Training Method)}} \\
NaVid~\cite{zhang2024navid}        & $0.95$\,M & $320$       & --   & --   \\
Uni-NaVid~\cite{zhang2024uninavid} & $3.6$\,M  & $1{,}400$   & --   & --   \\
NaVILA~\cite{cheng2024navila}      & $13.1$\,M & $1{,}152$   & --   & --   \\
NavFoM~\cite{zhang2025navfom}      & $8.0$\,M  & $4{,}032$   & --   & --   \\
OmniNav~\cite{xue2025omninav}      & $14.0$\,M & $14{,}592$  & --   & --   \\
ABot-N0~\cite{chen2026abotn0}      & $16.9$\,M & $6{,}144$   & --   & --   \\
\midrule
\rowcolor{cyan!15}\multicolumn{5}{c}{\textbf{Zero-Shot (Training-Free)}} \\
LaViRA-LA~\cite{ding2025lavira}  & $0$ & $0$ & $3.9$k$+$$0.2$k & $\sim$$1.65$P \\
LaViRA-VA~\cite{ding2025lavira}  & $0$ & $0$ & $1.0$k$+$$0.07$k & $\sim$$69$T \\
\textbf{\ourmethod-LA}           & \textbf{$0$} & \textbf{$0$} & $22.3$k$+$$1.6$k & $\sim$$9.6$P \\
\textbf{\ourmethod-VA}           & \textbf{$0$} & \textbf{$0$} & $0.99$k$+$$0.11$k & $\sim$$59$T \\
\bottomrule
\end{tabular}}
\end{table}

Beyond accessibility, the two regimes can be compared
head to head in raw FLOPs. NavFoM reports fine-tuning on 56 NVIDIA H100
GPUs for $\sim$$72$ hours, a total of $4{,}032$ H100-hours. At a
realistic $40\%$ MFU on H100 BF16 dense throughput of
$\sim$$989$\,TFLOPS\footnote{Per the NVIDIA H100 SXM5 datasheet, BF16
Tensor-Core throughput without sparsity.}, the training run consumes
$\sim$$5.7$\,ZFLOPs ($5.7\times 10^{21}$\,FLOPs), equivalent to roughly
$570$ full six-task \ourmethod evaluations (each $\sim$$10$\,EFLOPs).
A typical research project runs at most a few dozen full six-task
evaluations, well below this $570$ break-even, so in any realistic
budget \ourmethod is also cheaper in cumulative compute, not only more
accessible.

The per-call numbers in Table~\ref{tab:efficiency} come
from the same Full configuration as Section~\ref{subsec:main_results},
with Gemini-3.1-Pro for LA and Qwen3.5-27B for VA, measured over the
$1{,}800$-trajectory pool. Per call, LA averages $22{,}305$ input and $1{,}618$
output tokens, of which $88.7\%$ are reasoning-trace tokens, while VA
averages $\sim$$990$ input and $\sim$$110$ output tokens with no reasoning
trace. Per-call compute works out to $\sim$$9.6$\,PFLOPs for LA and
$\sim$$59$\,TFLOPs for VA, as listed in Table~\ref{tab:efficiency}. Of
the $\sim$$10$\,EFLOPs consumed by a six-task Full evaluation, $93.2\%$
goes to LA prefill, $6.0\%$ to LA reasoning tokens, $0.8\%$ to LA visible
output, and under $0.1\%$ to VA, so input-context length is the dominant
lever. In wall-clock terms, a complete six-task $600$-episode run finishes
in $5$--$7$ hours on $40$ parallel API workers, with a single LA call
taking $30$--$60$ seconds when the API is healthy and $80$--$150$ seconds
under contention, dominated by network round-trip and reasoning-token
decode, while the local planner and controller each run in under
$10$\,ms.

\subsection{Qualitative Analysis and Failure Modes}
\label{subsec:qualitative}

\begin{figure*}[!t]
  \centering
  \includegraphics[width=\linewidth,height=0.88\textheight,keepaspectratio]{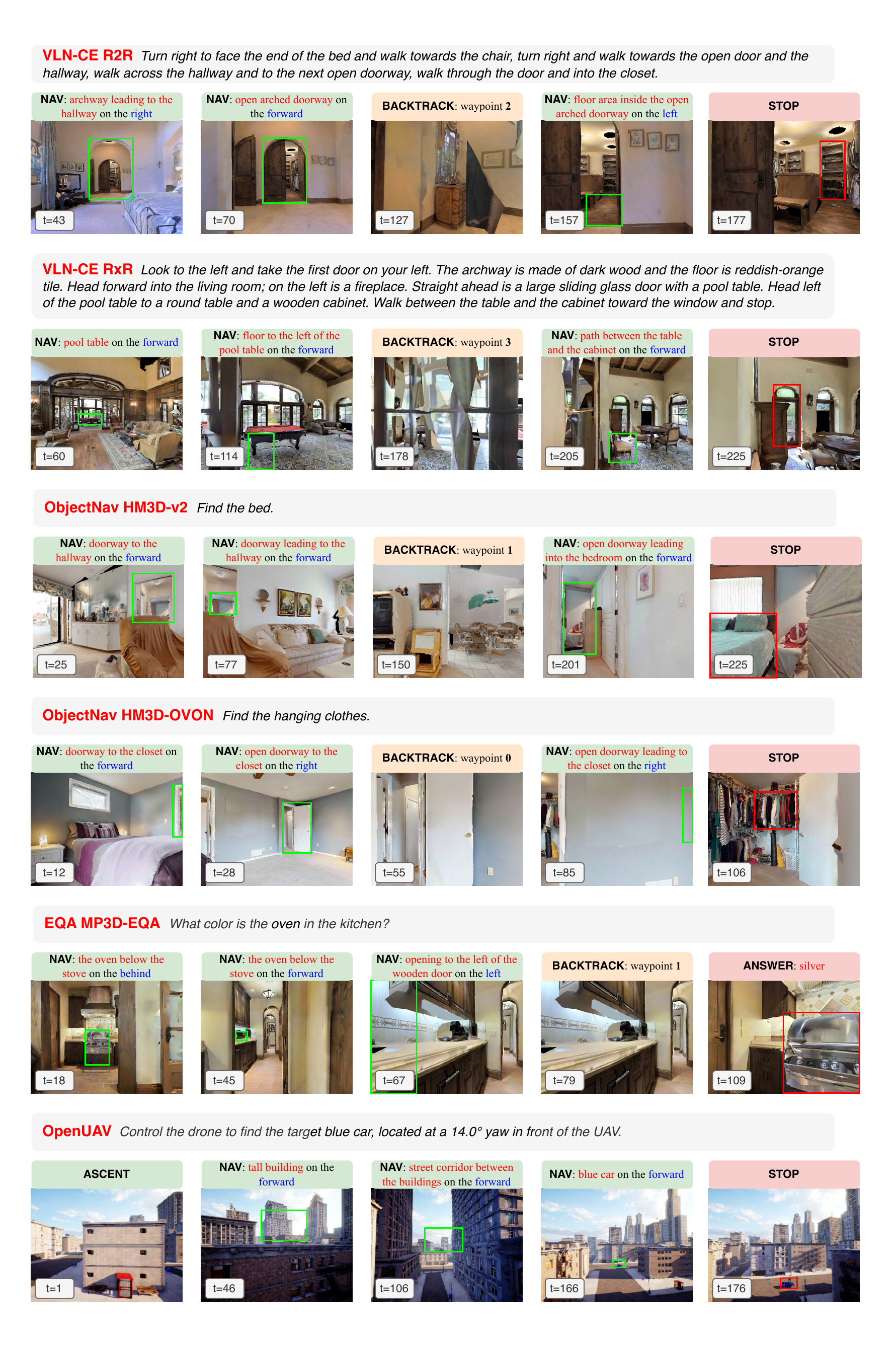}
  \caption{\textbf{Trajectory visualization across the
  six simulation benchmarks.}  Each row shows five keyframes from a
  representative successful episode, annotated with the literal Language
  Action emitted by the agent and the corresponding Vision Action
  grounding it on the first-person view.}
  \label{fig:sim_vis}
\end{figure*}

Figure~\ref{fig:sim_vis} shows two patterns in every row. Each \texttt{NAV}
keyframe is a grounded sub-goal phrase such as ``open arched doorway on the
forward'' or ``doorway leading to the hallway on the forward'', so the
Language Action Model commits in object- and region-level terms rather than as
raw waypoints. The explicit \texttt{BACKTRACK} action surfaces mid-trajectory
in five of six rows, including the EQA episode where the agent first considers
an oven, retreats to a previous waypoint, then locks the correct view and
answers ``silver''. This matches the SCB ablation in
Table~\ref{tab:ablation_mechanisms}. Backtracking is routine rather than a
rare recovery.

We pool all $1{,}800$ trials across the six benchmarks and classify failures
from the final-state metrics $\mathit{success}$, $\mathit{oracle\_success}$,
$\mathit{distance\_to\_goal}$, and $\mathit{path\_length}$, using a small set
of deterministic geometric rules. Table~\ref{tab:failure_modes} reports the
resulting categories.

\begin{table*}[t]
\centering
\caption{\textbf{Failure-mode breakdown across $1{,}800$ trials.}
Failures are categorised post-hoc from final-state
geometry, pooled across the six benchmarks. The top three categories
account for $\sim$$90\%$ of all failures; the two geometric tail modes
split the remainder.}
\label{tab:failure_modes}
\vspace{-6pt}
{\fontsize{9pt}{11pt}\selectfont
\setlength{\tabcolsep}{8pt}
\begin{tabular}{l|cccccc}
\toprule
\textbf{Failure Category} &
\textbf{$n$} & \textbf{\% all trials} & \textbf{\% of failures} &
\textbf{Dist.~to goal (m)} & \textbf{Path length (m)} & \textbf{Steps taken} \\
\midrule
Early stop on wrong target  & 319 & 17.7 & 45.8 & 11.0$\pm$4.73 & 11.9$\pm$6.9 & 217$\pm$107 \\
Reached but missed STOP     & 172 & 9.6  & 24.7 & 3.72$\pm$2.25 & 14.3$\pm$10.3 & 237$\pm$149 \\
Wrong answer (EQA)            & 136 & 7.6  & 19.5 & --   & --   & --   \\
Approached but undershot      & 48  & 2.7  & 6.9  & 4.13$\pm$0.57 & 9.6$\pm$7.3   & 170$\pm$119 \\
Wandered and lost               & 22  & 1.2  & 3.2  & 9.49$\pm$4.26 & 30.2$\pm$4.5  & 440$\pm$92  \\
\bottomrule
\end{tabular}}
\end{table*}

\textbf{Early stop on wrong target.} This mode accounts for $17.7\%$ of trials
and $45.8\%$ of failures. The agent triggers \texttt{STOP} after a short
trajectory of $\bar\ell{=}11.9$\,m and $\bar t{=}217$ steps, ending
$\sim$$11$\,m from the goal. The Language Action either misread the
instruction, locked onto the wrong landmark, or over-trusted a VA bounding box
on a distractor. This row dominates RxR at $34\%$ and R2R at $26\%$, where
long instructions make grounded landmarks easy to confuse.

\textbf{Reached but missed STOP.} This mode covers $9.6\%$ of trials and
$24.7\%$ of failures. $\mathit{oracle\_success}{=}1$ but the agent walks away
from the goal after passing the $3$\,m success ring. Trajectories are longer
at $\bar\ell{=}14.3$\,m and $\bar t{=}237$ steps, with a final distance of
$3.7$\,m. HM3D-OVON at $17\%$ and HM3D-v2 at $10\%$ are most affected,
consistent with their open-vocabulary targets being easier to over-shoot.

\textbf{Wrong answer on EQA.} This mode is $7.6\%$ of trials and $19.5\%$ of
failures, all on MP3D-EQA. Navigation may have been adequate but the LA's
final-view answer is wrong, a failure downstream of the Language backbone's
fine-grained visual reasoning.

\textbf{Geometric tail.} \emph{Approached but undershot} contributes $2.7\%$
of trials and \emph{Wandered and lost} $1.2\%$. Together they cover $10.0\%$ of
failures and under $4$ SR points. These are Robot-Action errors from depth
artefacts, FMM dead-ends, and infinite-loop traversals.

\section{Real-World Deployment}
\label{sec:realworld}

\begin{figure*}[!t]
  \centering
  \includegraphics[width=\linewidth,height=0.6\textheight,keepaspectratio]{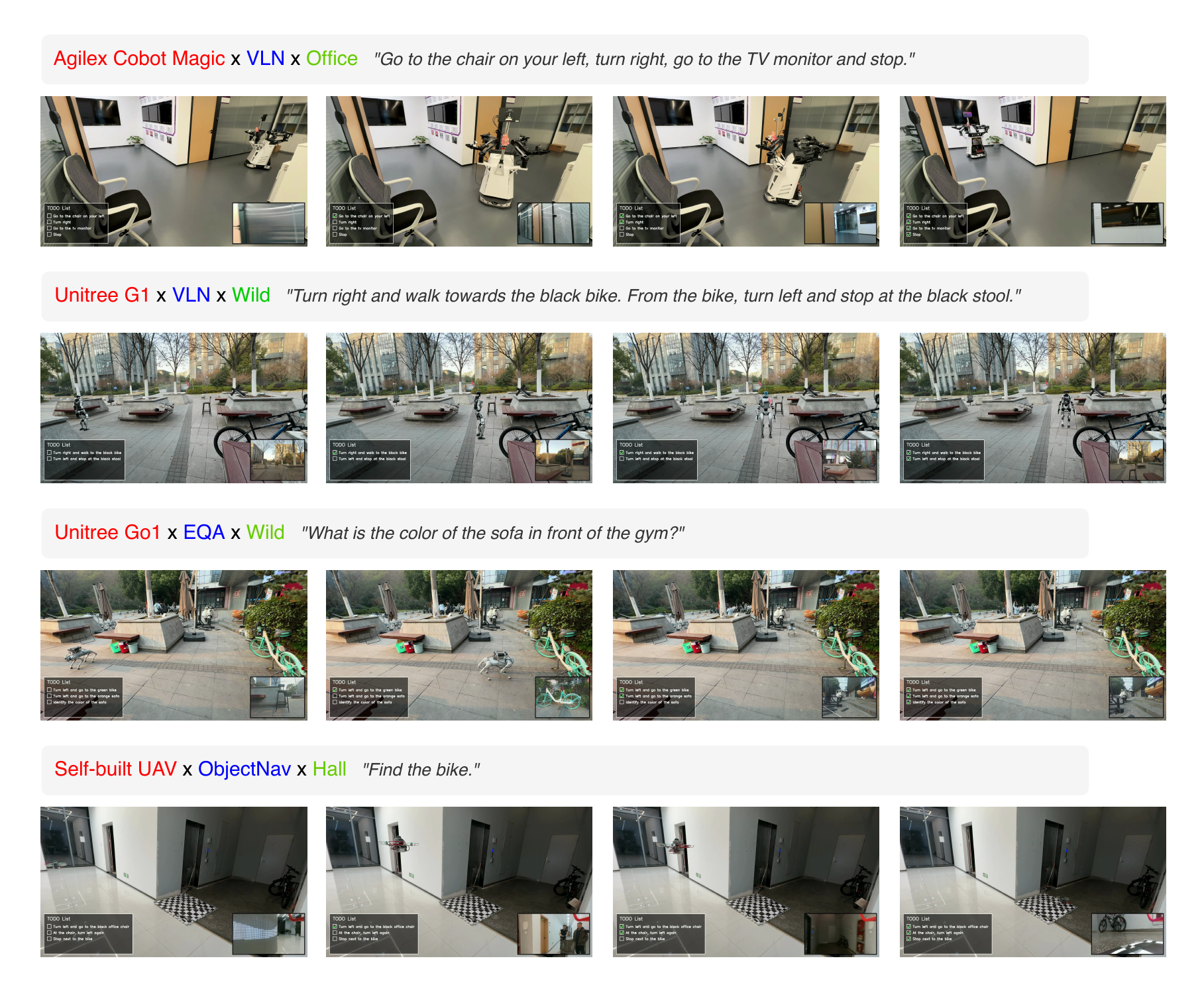}
  \caption{\textbf{Real-world deployment across four
  embodiments, three task types, and three environments.} An Agilex Cobot
  Magic bimanual wheeled platform, a Unitree~G1 humanoid, a Unitree~Go1
  quadruped, and a self-built quadrotor UAV share byte-identical Language
  and Vision Action Models across the three ground robots; only the
  low-level controller and the inference target change per platform.}
  \label{fig:realrobots}
\end{figure*}

We deploy \ourmethod on four heterogeneous embodiments shown in
Figure~\ref{fig:realrobots}, an Agilex Cobot Magic bimanual wheeled platform,
a Unitree~G1 humanoid, a Unitree~Go1 quadruped, and a self-built quadrotor
UAV. These four robots span three task families: VLN-CE, EQA, and ObjectNav.
They also span three environment types: indoor office, indoor hall, and
outdoor wild. No per-platform retraining is performed. The Language and Vision
Action Models are byte-identical across the three ground robots, and only the
low-level controller and the inference target change per platform. Ground
robots run the MLLMs locally, while the UAV routes both calls through a remote
API stack because its thermal and weight budgets rule out on-board inference
at the required capacity.


\subsection{Per-Embodiment Setup}

For the three ground robots we replace the simulation-time FMM planner with
NavDP~\cite{cai2025navdp} as the Robot Action stage. NavDP turns the current
RGB-D observation into a short trajectory in the robot frame, which the
platform's native velocity controller tracks. The UAV uses the PX4
position-control loop instead, because flight requires three-dimensional
waypoint commands rather than planar velocity. Sensor calibration is the only
platform-specific tuning step, and a hand--eye solve plus a checkerboard
alignment of the body-fixed cameras converge in under fifteen minutes per
platform.

\textbf{Agilex Cobot Magic (wheeled bimanual).} A chest-mounted Intel
RealSense D435i RGB-D camera sits on a 2-DoF differential base, inherited from
LaViRA~\cite{ding2025lavira}. NavDP trajectories are tracked by a proportional
velocity controller. An onboard RTX~4090 serves a single Qwen3.5-27B-Q4 model
for both LA and VA with no remote API call during operation.

\textbf{Unitree G1 (humanoid).} Four Orbbec Gemini 336L RGB-D cameras
supply the four-view panorama, NavDP trajectories are executed by the native
locomotion controller at $(v_x,v_y,\omega_z)$ sampled at $1$\,Hz with no gait
re-training, and an onboard Jetson Orin NX runs Qwen3.5-9B-Q4 for both LA and
VA.

\textbf{Unitree Go1 (quadruped).} Four Orbbec Gemini 336L
RGB-D cameras supply the four-view panorama, sharing the same Jetson Orin
NX and Qwen3.5-9B-Q4 stack as the Unitree G1. NavDP
trajectories are tracked by the native velocity-mode controller. The quadruped
is the most agile of the four platforms and executes wide turns without
re-planning.

\textbf{Self-built UAV (quadrotor).} An Intel RealSense D435i matching the
OpenUAV simulation provides RGB-D, and a Jetson Orin NX handles sensor I/O.
The PX4 controller consumes $(x,y,z)$ waypoint commands produced by NavDP from
the depth observation and the target waypoint. LA runs on Gemini-3.1-Pro and
VA on Qwen3.5-27B through a remote API stack. The single largest engineering
surprise here was network jitter. A single LA call takes $30$--$60$\,s when
the API is healthy and up to $150$\,s under contention, so we extend the plan
horizon to roughly $5$\,m per call. Shorter horizons leave the UAV in a stale
plan by the time the next observation lands, which we saw as occasional
hover-and-redecide loops on the first deployment day.

\subsection{Trials and Engineering Cost}

We evaluate in an indoor office of $\sim$$400$\,m$^2$, an outdoor corridor
of $60$\,m, and an outdoor wild scene
(Fig.~\ref{fig:realrobots}). Tasks cover three simulation families with example instructions
``Go to the chair on your left, turn right, go to the TV monitor and stop.''
for instruction following, ``Find the bike.'' for open-vocabulary object
finding, and ``What is the color of the sofa in front of the gym?'' for EQA.
Real-world rollouts are run as qualitative validation rather than a
quantitative benchmark, since simulation already provides a controlled setting
for systematic metrics. UAV failures are dominated by altitude-induced visual
ambiguity where the agent reaches the goal vicinity but cannot confirm the
target from above and circles instead. The same verbatim panoramic prompt
drives the Agilex Cobot Magic, Unitree G1, and the Unitree Go1, without edits, which we read as
direct evidence that the decomposition abstracts away the viewpoint.

\textbf{Engineering effort per embodiment.} Adding a new
embodiment takes only sensor calibration in $\sim$$2$\,h, a
velocity-controller adapter that maps 2-D waypoints to the platform's native
command interface in $\sim$$4$--$6$\,h with a 3-D extension for the UAV, and a
brief integration test in $\sim$$2$\,h. Total engineering effort across all
four embodiments was roughly 40 person-hours, against the hundreds of
GPU-hours typically required to retrain a VLA per platform. The Language and
Vision Action Models are never modified.

\textbf{Field-only failure modes.} Three failure modes appear in the real
world but not in simulation. \emph{Lighting drift} shifts Vision Action
confidences enough that target colours occasionally flip on object-finding
episodes, mitigated by an exposure-locked frame at the start of each episode.
\emph{Network jitter} on the UAV triggers PX4's safety hover whenever a single
LA call exceeds $150$\,s, costing roughly one extra step per affected episode.
\emph{Calibration drift} of $\sim$$2$\,cm after thirty
minutes of walking makes the Unitree~G1 take overly wide turns at narrow
doorways, fixed by a quick recalibration between task batches.

\section{Conclusion}
\label{sec:conclusion}

We presented \ourmethod, a training-free agentic framework that factors
embodied navigation into three levels. The Language Action Model emits
semantic-level directional commands, the Vision Action Model emits pixel-level
visual targets, and the Robot Action controller back-projects those targets
into the robot's native action space. Two agent-loop mechanisms make this
unification practical. TODO List Memory keeps the agent's plan explicit by
rewriting a checklist of pending sub-goals at every step. Second Chance
Backtrack rolls the robot back to its pre-error state and conditions the next
plan on the failed sub-trajectory.

A single zero-shot implementation reaches $60.7\%$ SR on VLN-CE R2R, $51.3\%$
on VLN-CE RxR, $77.7\%$ on HM3D-v2, $60.0\%$ on HM3D-OVON, $54.7\%$ on
MP3D-EQA, and $40.00\%$ on OpenUAV, matching or even surpassing trained
foundation models that consume millions of samples and thousands of GPU-hours. The same agentic core
deploys to four heterogeneous real robots, namely an Agilex Cobot~Magic
bimanual wheeled platform, a Unitree~Go1 quadruped, a Unitree~G1 humanoid, and
a self-built quadrotor UAV, by swapping only the low-level controller.

\textbf{Limitation and Future Work.} Four limitations remain that map
naturally onto the next steps. (i)~The strongest backbones such as
Gemini-3.1-Pro are proprietary, and open-weight alternatives still trail by a
clear margin. We will distil agent traces from the proprietary Language and
Vision Action Models into open backbones so the full pipeline can be
self-hosted. (ii)~Large-area grounding is unreliable, since ``the hallway'' is
harder to localise than ``a sofa''. We will let the Vision Action Model call
SAM~\cite{kirillov2023sam} and Grounding~DINO~\cite{liu2024grounding} as
auxiliary tools when its bounding-box confidence is low, rather than as
replacements. (iii)~Long-instruction tasks such as RxR still leave a clear gap
to trained VLA models, the dominant remaining limitation of a training-free design.
We will extend TDM with hierarchical sub-goal compression and longer-context
backbones to sustain attention over hundreds of words. (iv)~Dynamic obstacles
and humans are handled only by reactive low-level control. Lifting
pedestrian-intent reasoning into the Language Action Model is the path to true
social navigation, and is a direct extension of the TDM and SCB loop rather
than a new architecture.

{\fontsize{6pt}{7pt}\selectfont
\bibliographystyle{IEEEtran}
\bibliography{IEEEabrv,references}}

\end{document}